\documentclass{article} 
\usepackage{iclr2025_conference,times}

\usepackage{amsmath,amsfonts,bm}

\def\eqref#1{equation~\ref{#1}}

\def\1{\bm{1}}

\DeclareMathAlphabet{\mathsfit}{\encodingdefault}{\sfdefault}{m}{sl}
\SetMathAlphabet{\mathsfit}{bold}{\encodingdefault}{\sfdefault}{bx}{n}

\usepackage{tabularx}
\usepackage{url}
\usepackage{graphicx}
\usepackage{booktabs}
\usepackage{algorithm}
\usepackage{xcolor}
\usepackage{wasysym}
\usepackage{amssymb}
\usepackage{algpseudocode}
\usepackage{caption}
\usepackage{subcaption}
	
\usepackage{color, colortbl}
\iclrfinalcopy

\usepackage[accsupp]{axessibility}  
\usepackage[pagebackref,breaklinks,colorlinks]{hyperref}
\definecolor{Type1Blue}{HTML}{1f77b4}
\definecolor{Type2Orange}{HTML}{ff7f0e}
\definecolor{Type3Green}{HTML}{2ca02c}
\definecolor{Type4Red}{HTML}{d62728}
\definecolor{Type5Purple}{HTML}{9467bd}
\definecolor{Type6Brown}{HTML}{8c564b}
\usepackage{pifont}  

\definecolor{LightCyan}{rgb}{0.88,1,1}
\definecolor{Gray}{gray}{0.9}
\title{LIME-Eval: Rethinking Low-light Image Enhancement Evaluation via Object Detection}

\author{Mingjia Li, Hao Zhao, and Xiaojie Guo\thanks{Corresponding Author.} \\
College of Intelligence and Computing,
Tianjin University, China\\
\texttt{\{mingjiali,student\_zh\}@tju.edu.cn, xj.max.guo@gmail.com}
}

\begin{document}

\maketitle

\begin{abstract}

Due to the nature of enhancement--the absence of paired ground-truth information, high-level vision tasks have been recently employed to evaluate the performance of low-light image enhancement.  A widely-used manner is to see how accurately an object detector trained on enhanced low-light images by different candidates can perform with respect to annotated semantic labels. In this paper, we first demonstrate that the mentioned approach is generally prone to overfitting, and thus diminishes its measurement reliability. In search of a proper evaluation metric, we propose LIME-Bench, the first online benchmark platform designed to collect human preferences for low-light enhancement, providing a valuable dataset for validating the correlation between human perception and automated evaluation metrics. We then customize LIME-Eval, a novel evaluation framework that utilizes detectors pre-trained on standard-lighting datasets without object annotations, to judge the quality of enhanced images. By adopting an energy-based strategy to assess the accuracy of output confidence maps, our LIME-Eval can simultaneously bypass biases associated with retraining detectors and circumvent the reliance on annotations for dim images. Comprehensive experiments are provided to reveal the effectiveness of our LIME-Eval. Our benchmark \href{https://huggingface.co/spaces/lime-j/eval}{platform} and \href{https://github.com/lime-j/lime-eval}{code} are available online.

\end{abstract}

\section{Introduction}
Low-light conditions significantly challenge imaging by reducing the visibility of important details and/or introducing distortions in captured images, such as noise, blur, and color shifts. The poor quality of images captured in such conditions not only hampers everyday photography but also poses serious issues in fields where image clarity is critical, such as surveillance, navigation, and astrophotography. Consequently, low-light image enhancement has emerged as an essential technique to improve the quality of images taken in unsatisfactory lighting environments.

The objective of low-light image enhancement is to concurrently  brighten dark regions, enhance suppressed details, preserve color fidelity, and eliminate potential artifacts. In other words, it is desired to generate high-quality images that closely resemble those taken under ``good"\footnote{As an enhancement task, there is no well-defined optimal lighting condition.} lighting conditions. While substantial progress has been made in this domain, spanning from histogram equalization~\citep{DBLP:conf/icpr/TrahaniasV92} to advanced deep-learning approaches~\citep{DBLP:conf/iccv/CaiBLWTZ23, DBLP:conf/mm/ZhangZG19, DBLP:journals/ijcv/GuoH23}, a key challenge persists in objectively evaluating the performance of enhancement algorithms. Historically, the image quality assessment (IQA) of enhancement has depended on reference-based metrics (\emph{e.g.}, PSNR and SSIM), which compare the enhanced results to a reference image deemed to be of high quality. 

We argue that reference-based metrics are unsuitable for low-light \textbf{enhancement}, because the very nature of the problem precludes the existence of reliable reference images. On the one hand, capturing such references with identical settings, except for proper illumination, is inherently challenging. On the other hand, even if reference images are obtained under well-controlled settings, many/infinite variations of ``well-lit" conditions exist, making it hard to determine which specific scenario aligns with the ``best". This absence of a definitive standard (actually for all enhancement tasks) complicates the evaluation process, and necessitates the development of no-reference assessments. One might suggest using no-reference IQA metrics, such as NIQE~\citep{Mittal2013Making} and BRISQUE~\citep{DBLP:journals/tip/MittalMB12}, which are currently mainstream. However, assessing the quality of enhanced images involves a complex interplay of quality and aesthetic issues (color restoration and lightness). Existing no-reference IQA metrics are often inconsistent or even contradictory with human perception, rendering them unreliable in low-light scenarios~\citep{DBLP:conf/cvpr/SahaMB23, DBLP:journals/ijcv/GuoH23}, as we will demonstrate in Sec.~\ref{sec:low}.

Alternatively, the community has begun to reassess the effectiveness of low-light image enhancement techniques by examining their impact on downstream  vision tasks, particularly object detection. The core idea is that the performance of downstream models can serve as a proxy for human perception. If objects within an image are adequately illuminated, they should be easily identifiable by both humans and machines. A widely-adopted evaluation protocol in recent studies requires fine-tuning an object detector on images enhanced by different methods~\citep{cui2022need, cai2023retinexformer, han2024glare}.
However, this retraining process raises a significant concern: \emph{i.e.}, overfitting. The detector may become overly tailored to the specific characteristics of the enhanced images, disregarding its resemblance to a natural, well-illuminated image. This leads to a fundamental question: \emph{Is fine-tuning detectors a valid approach to evaluating enhancers?} In Sec.~\ref{sec:low}, we will demonstrate that the performance of fine-tuned detectors does not necessarily correlate with the quality of enhanced images. With the application of an appropriate augmentation strategy, fine-tuning on inadequately enhanced images can still yield better detection performance than even the most advanced enhancement methods. This finding indicates that the fine-tuning protocol conflates the effectiveness of the enhancement algorithms with the adaptability of detection models, ultimately compromising the fairness and reliability of the evaluation. 

To remedy the aforementioned flaw, a straightforward strategy shall deploy detectors pre-trained on data captured under normal-lighting conditions to evaluate enhanced images, using annotations from the original low-light images~\citep{DBLP:conf/cvpr/WangY021, DBLP:conf/cvpr/SCI}. The underlying premise is that the closer the enhanced results resemble the normal-lit image domain, the better the detection performance will be.
To validate this approach, we introduce the first online benchmark platform, \emph{LIME-Bench}, designed to collect human preference on assessing low-light enhancement methods. Through the data collected on this platform, we verify that directly applying pre-trained detectors serves as a more effective critique for evaluating low-light image quality than a series of previously applied quality assessment methods, offering a reliable proxy for enhancement performance.

While this manner takes advantage of the inherent generalizability of models trained under standard illumination to assess the fidelity of low-light enhancements, it introduces its own set of challenges related to semantic labels. For one thing, obtaining accurate annotations for low-light images is more difficult and time-consuming than for those captured under normal lighting conditions, as the reduced visibility and contrast in low-light images increase ambiguities in object boundaries and classifications. For another thing, the reliance on annotated labels restricts the flexibility of evaluation. Moreover, annotating a large dataset of low-light images to establish a reliable benchmark for evaluation is labor-intensive, limiting the scalability of this approach.

This work proposes a novel framework, called \emph{LIME-Eval}, for evaluating low-light image enhancement. Grounded in a pioneering energy-based criterion, our method sidesteps the biases and time-consuming processes associated with retraining detectors while liberating the demand for both reference images and detection labels. These features broaden the applicability of LIME-Eval to unlabeled and reference-free low-light scenarios. 
Our primary contributions are summarized as follows:
\begin{itemize} 
    \item By retraining detectors on enhanced images produced by various low-light enhancement methods, we find that, under appropriate data augmentation conditions, higher detection accuracy does NOT necessarily correlate with superior enhancement quality. 
    \item We collect 6,362 feedback pairs from 750 users, encompassing factors such as blurriness, exposure, noisiness, color, and overall quality across 14 low-light enhancement methods to construct the first low-light user preference dataset, LIME-Bench. Utilizing this data, we benchmark non-reference image quality assessment methods in prior arts and validate the correlation between detector-based evaluations and human preferences.
    \item We introduce a novel energy-based evaluation framework, say LIME-Eval, which effectively links the quality of enhanced images with the performance of object detection without object labels or reference images. Comprehensive experimental results and analyses confirm LIME-Eval's effectiveness and evidence its potential to guide low-light  enhancers. 
\end{itemize}
\section{Related Work}

\noindent\textbf{Low-light Image Enhancement} 
aims to tackle multiple degradations present in dark images such as noise, low contrast, and color shift. Early methods, like histogram equalization and its variants, sought to improve image visibility by adjusting global and/or local contrast. The advent of deep learning has led to a range of innovative approaches. Within this context, Retinex theory--which conceptualizes an image as the product of reflectance and illumination components--has gained significant attention. Several schemes based on this paradigm endeavor to produce normal-light images by modulating the illumination component and estimating reflectance~\citep{Fu_2016_CVPR, doi:10.1137/100806588}. In advancing the exploration of attention mechanisms, the transformer architecture integrates self-attention with convolutional processes to simultaneously extract long-range and short-range dependencies. As a representative work, Retinexformer~\citep{cai2023retinexformer} introduces a new self-attention IG-MSA module, based on retinex theory and transformer architecture. Focusing on the illumination of an image, an illumination adaptive transformer (IAT) was proposed~\citep{cui2022need}, notable for its minimalist design of just 90k parameters and its efficiency in addressing illumination adjustments. Guo and Hu~\citep{DBLP:journals/ijcv/GuoH23} decoupled the entanglement of noise and color distortion, further alleviating the challenges of low-light enhancement in the presence of complex degradations. In the absence of ground truth, Guo \emph{et al.}~\citep{Guo_2020_CVPR} proposed an unsupervised method adjusting the illumination with LE-curve, achieving reasonable results at an impressively fast pace. These developments represent a significant leap in low-light image enhancement. However, as previously discussed, the lack of exact reference images for enhancement tasks necessitates further research to explore methods for assessing enhanced images in reference-free fashions.

\noindent\textbf{Image Quality/Aesthetic Assessment} 
has always been a fundamental task in image processing, especially in enhancement, compression, and restoration. Traditional methods heavily rely on full-reference metrics, with Peak Signal-to-Noise Ratio (PSNR) and Structural Similarity Index (SSIM) as two prominent examples, comparing processed results against high-quality references. Recently, several deep-learning-based variants for full-reference image quality assessment have been proposed. For instance, Kim \emph{et al.}~\citep{8970465} employed an end-to-end CNN model. While being effective under certain conditions, this approach frequently fails to encompass the perceptual quality as perceived by human observers. This discrepancy has spurred the development of no-reference image quality assessment (NR-IQA) methods, which forgo the need for any reference images. Early NR-IQA research primarily focused on specific distortions, notably JPEG compression~\citep{1038064, MARZILIANO2004163}. The introduction of the LIVE dataset~\citep{DBLP:journals/tip/SheikhSB06} marked a shift toward general-purpose NR-IQA, which leverages natural scene statistics (NSS) from spatial~\citep{6272356, 6353522} or transform domains~\citep{5756237, 6172573} to assess image quality, predicated on the premise that deviations from the statistical regularities found in natural images correlate with perceived visual quality.~\citep{doi:10.1146/annurev.neuro.24.1.1193}. With the expansion of IQA datasets and the growing influx of images, deep learning has emerged as the predominant force in NR-IQA. To compensate for the shortage of manually-labeled data, strategies like patchwise training~\citep{8063957, Kang_2014_CVPR}, transfer learning~\citep{8451285}, and quality-aware pre-training~\citep{8110690, Liu_2017_ICCV} have been developed. Up-to-date NR-IQA research cooperated with innovations like active learning, meta-learning, patch-to-picture mapping, loss normalization, and adaptive convolution. These advancements aim to enhance generalizability~\citep{9399249}, enable rapid adaptation~\citep{Zhu_2020_CVPR}, improve local quality prediction~\citep{Ying_2020_CVPR}, expedite convergence~\citep{10.1145/3394171.3413804}, and facilitate content-aware quality assessment~\citep{Su_2020_CVPR}. These IQA methods have demonstrated significant success in evaluating the quality of enhanced images. Nevertheless, there remains a noticeable gap in understanding the relationship between these quality evaluations and the performance of subsequent downstream tasks. Bridging this gap is crucial for a more comprehensive assessment of image enhancement techniques in real-world applications.

\noindent\textbf{Benchmarking Low-light Enhancement with Detection} is tough due to the subjective nature of image quality and the lack of suitable standards for comparison. This has led researchers to explore alternative evaluation strategies, \emph{e.g.}, subjective assessment by human observers, or the use of synthetic datasets where ground truth is artificially generated. The ExDark dataset~\citep{LOH201930} serves as a repository of low-light object images, defined by criteria including low illumination levels or pronounced variations in lighting. The Matching in the Dark (MID) dataset~\citep{Song_2021_ICCV} offers a collection of stereo image pairs spanning 54 indoor and 54 outdoor settings, culminating in a total of 108 scene pairs. Despite the value of these datasets, they suffer from limitations in scalability and often fail to capture the complexity of real-world scenarios. Thus, it is imperative to develop innovative methodologies for evaluating enhanced images in the absence of reference images.

\noindent\textbf{Energy-based Models} (EBMs) are versatile, non-normalized probabilistic models introduced by~\citep{lecun2006tutorial}. They define relationships among variables by assigning a scalar energy value to each multivariate instance. Unconstrained by the need to maintain normalized probabilities, EBMs have found application across a wide array of tasks~\citep{DBLP:conf/collas/LiDVM22, DBLP:conf/nips/DuLM20, DBLP:conf/icml/DuLTM22}. Thanks to their ability to represent complex, high-dimensional data distributions, EBMs have also been applied in generative modeling tasks~\citep{DBLP:conf/iclr/ArbelZG21}. The work~\citep{DBLP:conf/iclr/GrathwohlWJD0S20} demonstrates how classifiers can inherently function as EBMs, further broadening their applicability. This perspective on energy has been harnessed for tasks such as out-of-distribution detection~\citep{DBLP:conf/icml/LafonRRT23} and automated evaluation of classification models~\citep{peng2024energy}. Inspired by these advancements, our approach adopts energy-based statistics as a proxy for average accuracy, showcasing the model's adaptability and effectiveness in evaluation contexts.

\label{sec:overf}

\section{Rethinking Evaluation Protocol}
\label{sec:low}
The information bottleneck theory~\citep{DBLP:journals/corr/physics-0004057} suggests that neural network operations can result in information loss, potentially obscuring critical clues for high-level tasks. This understanding has influenced the evaluation of low-light enhancement methods, where performance is often assessed by retraining downstream recognition models on images enhanced by these enhancers. But, the validity of such an evaluation scheme is questionable. To manifest this, we carefully select four low-light enhancement techniques that exemplify the diversity in current low-light enhancement approaches, including Zero-DCE~\citep{Guo_2020_CVPR}, Bread~\citep{DBLP:journals/ijcv/GuoH23}, IAT~\citep{cui2022need} and RetinexFormer~\citep{cai2023retinexformer}.

\begin{table}[t]
    \centering
    \caption{Quantitative comparison in detection accuracy mAP on the test split of ExDark. The best and second-best results by each scheme are in \textbf{bold} and \underline{underlined}, respecitvely. The results in the `Dim image' column are obtained by directly training the detector on images without enhancement. }
    \begin{tabular}{c|c|c|c|c|c|c}
    \hline
        & Dim image & Zero-DCE & Bread & Bread-round& IAT & RetinexFormer \\
        \hline
        Enhance First & N/A  & \underline{45.9} & \underline{45.9} & N/A & \textbf{47.6} &  46.1\\
        \rowcolor{Gray}
        Augmentation First  & \textbf{49.2} & \underline{49.2} & 48.9 & 48.8& 49.0 & 48.5 \\
        Direct Eval & 35.0 & 34.0 & \textbf{35.4} & N/A & \underline{35.3} & 34.2 \\
    \hline
    \end{tabular}
    
    \label{tab:boost_with_aug}
\end{table}
\begin{figure}[t]
    \centering
    
    \includegraphics[width=.19\linewidth]{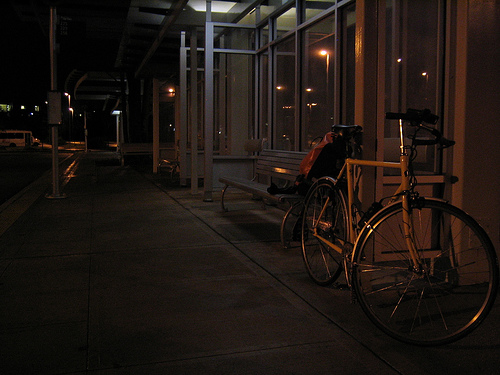}
    \includegraphics[width=.19\linewidth]{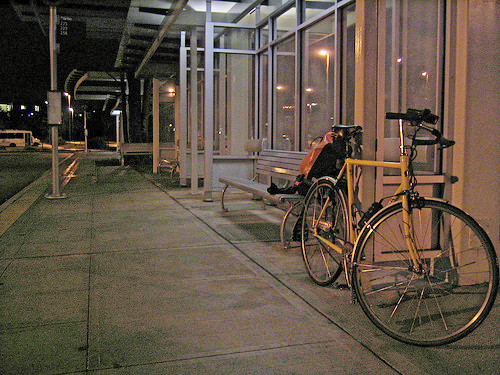}
    \includegraphics[width=.19\linewidth]{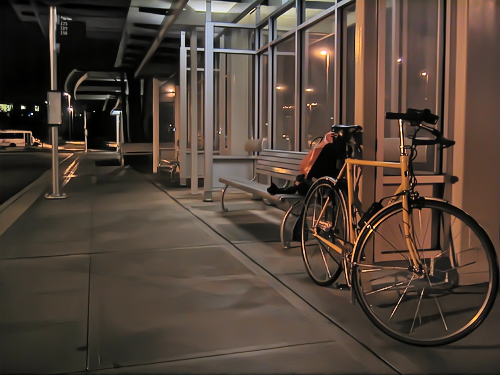}
    \includegraphics[width=.19\linewidth]{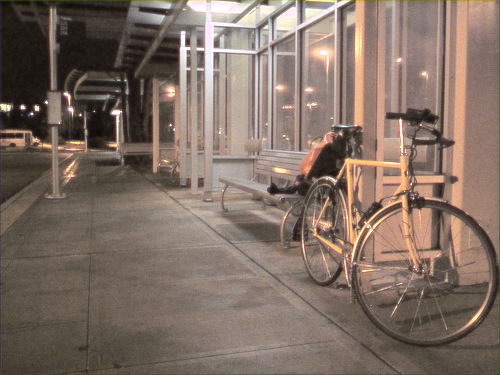}
    \includegraphics[width=.19\linewidth]{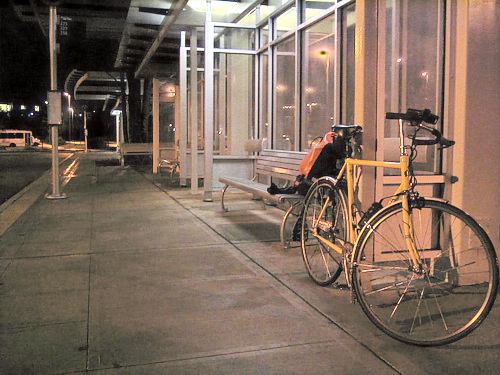}

    \includegraphics[width=.19\linewidth]{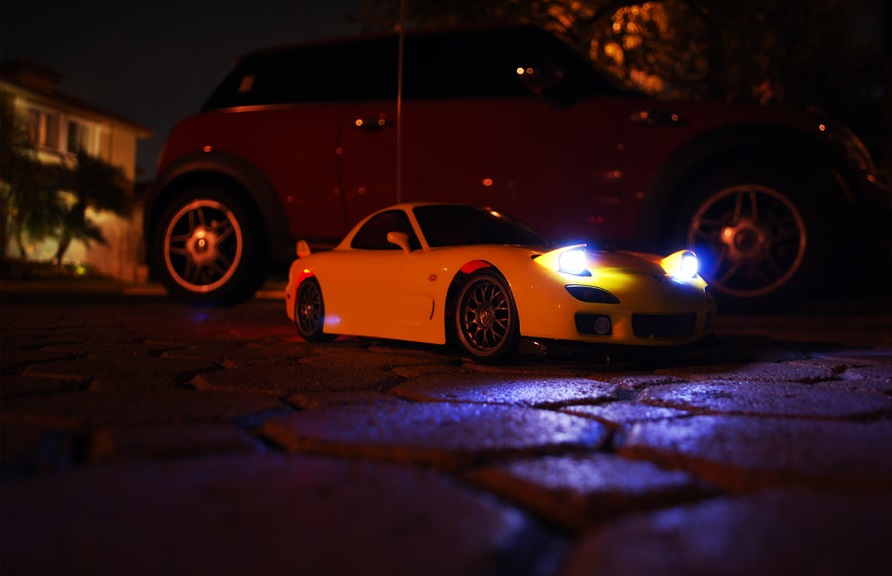}
    \includegraphics[width=.19\linewidth]{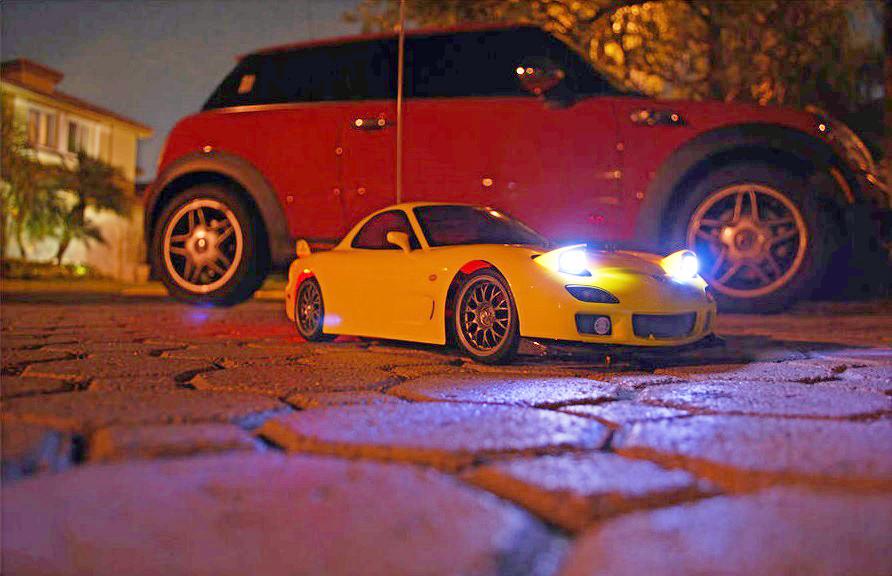}
    \includegraphics[width=.19\linewidth]{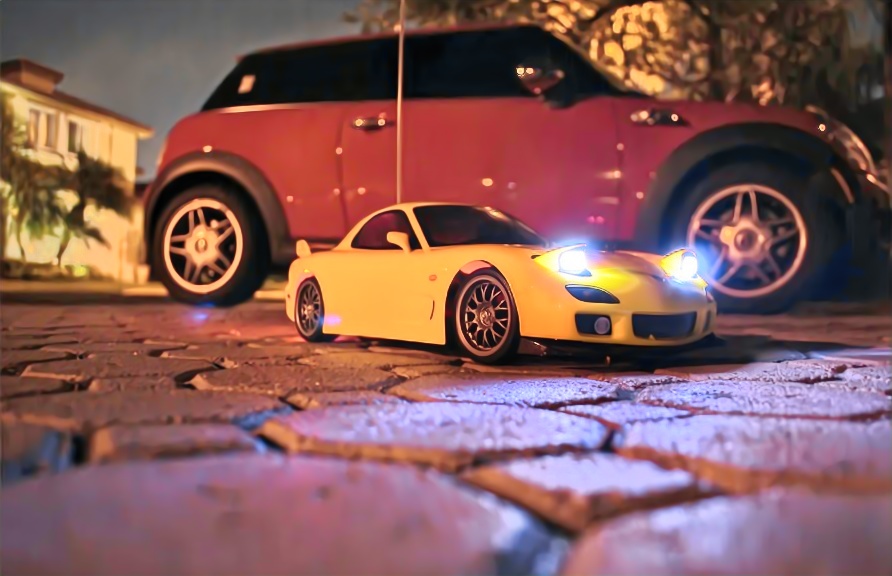}
    \includegraphics[width=.19\linewidth]{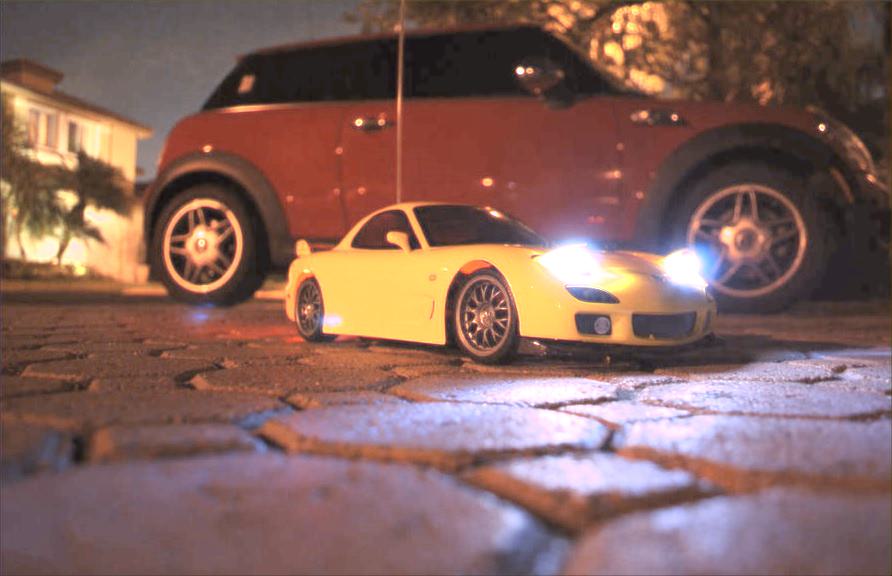}
    \includegraphics[width=.19\linewidth]{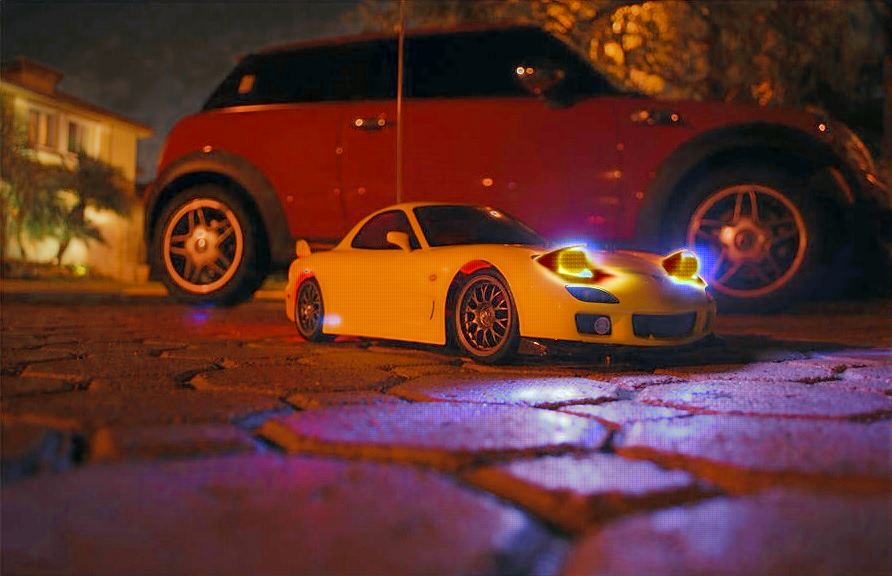}
    
    \includegraphics[width=.19\linewidth]{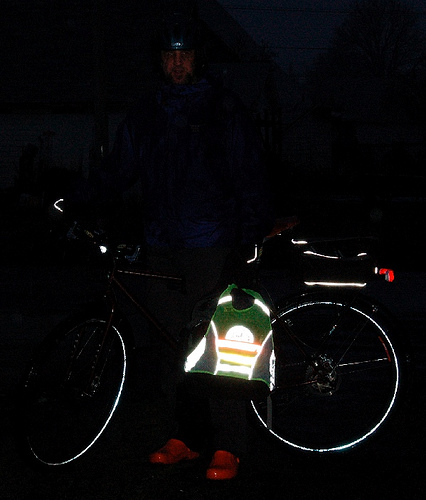}
    \includegraphics[width=.19\linewidth]{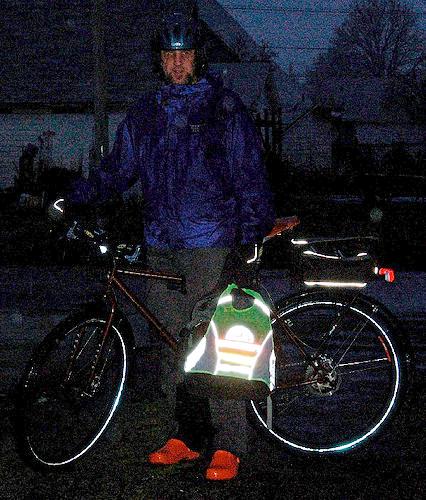}
    \includegraphics[width=.19\linewidth]{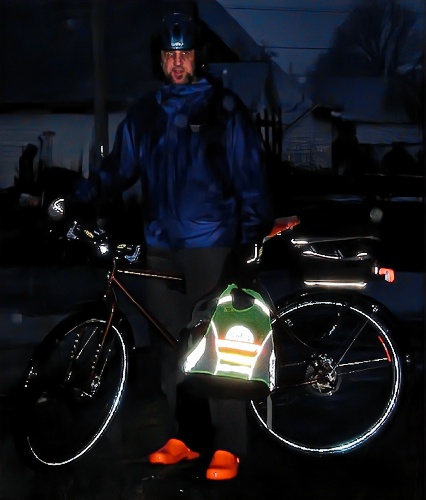}
    \includegraphics[width=.19\linewidth]{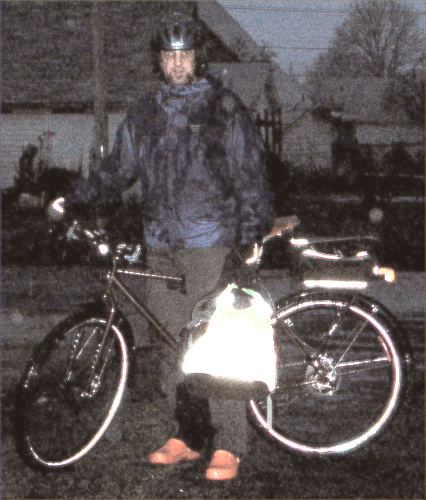}
    \includegraphics[width=.19\linewidth]{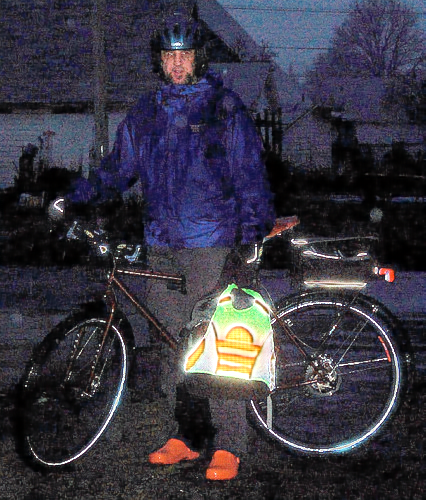}

    \begin{subfigure}{0.19\linewidth}
        \centering
        \subcaption{Input}
    \end{subfigure}
    \begin{subfigure}{0.19\linewidth}
        \centering
        \subcaption{Zero-DCE}
    \end{subfigure}
    \begin{subfigure}{0.19\linewidth}
        \centering
        \subcaption{Bread}
    \end{subfigure}
    \begin{subfigure}{0.19\linewidth}
        \centering
        \subcaption{IAT}
    \end{subfigure}
    \begin{subfigure}{0.19\linewidth}
        \centering
        \subcaption{RetinexFormer}
    \end{subfigure}
    \caption{Qualitative comparisons on samples from the ExDark dataset. Please zoom in for more details. Due to the page limit, more cases can be found in the appendix.}
    \label{fig:quali_ExDark}
\end{figure}

Without loss of generality, we initiate our investigation with object detection on the ExDark dataset, a widely recognized benchmark for low-light conditions. As for our base detector, we choose the medium version of YOLOX, aiming to provide a balanced and comprehensive analysis of the performance of low-light enhancers in challenging real-world conditions. All the involved enhancers, pre-trained on the LOL dataset, remain fixed during detector training. We experiment with two distinct training settings: 
1) \emph{Enhancement First Augmentation After}: Low-light images are enhanced first and saved in a lossless format, strong augmentation techniques (\emph{e.g.}, Mosaic, shear, mixup); and 2) \emph{Augmentation First Enhancement After}: Augmentation techniques are applied directly to the original low-light images, enhancement is performed on-the-fly, applied to the augmented output.

Referencing Tab.~\ref{tab:boost_with_aug}, it is noteworthy that while Bread exhibits superior visual results (please see Fig.~\ref{fig:quali_ExDark} for visual comparisons), it does not achieve the highest detection performance under either of the tested settings. Interestingly, the detection performance of Bread (45.9), IAT (47.6), and RetinexFormer (46.1), when enhanced before data augmentation, does not surpass dim (low-light) images (49.2). In contrast, when adopting the augmentation-first strategy, all the methods show a marked improvement in performance. To ensure that performance gains is not attributable to information loss during quantization, we conducted an experiment with Bread by clipping and quantizing its intermediate output before detector training. The modified version, referred to as Bread-round, demonstrated performance comparable to the original Bread. This indicates that the performance drop in the enhancement-first setting is inherent to the enhancement-first scheme itself.

This evidence highlights a key limitation of the retraining approach: it encourages detectors to optimize the utilization of available input clues and adapt specifically to the enhanced input domain. Under this paradigm, the training process compels the detector to rely solely on these input clues, rather than leveraging the common sense that underpins human perception.
To tackle the overfitting issue, a straightforward solution is to forego training the detector, and instead use models trained on large-scale normal-light datasets (for instance, the MS-COCO dataset) for direct inference on low-light images.  The results is detailed in Tab.~\ref{tab:boost_with_aug}, under ``Direct Eval". The findings reflect that Bread and IAT, which visually resemble normal-light images more closely, outperform those without enhancement and models Zero-DCE (which suffers from poor noise suppression) and RetinexFormer (which introduces artifacts due to overfitting). As the misalignment between the focus of fine-tuned detectors and actual perceptual quality results in skewed evaluation fairness, direct evaluation using pre-trained detectors without additional fine-tuning seems to offer a more unbiased protocol for assessing low-light enhancement methods. 

\section{Benchmarking Machine-Human Consensus via LIME-Bench} %
\label{sec:low}
However, the direct evaluation approach raises new questions that warrant further investigation: \emph{Is there always a consensus between human preference and direct detection performance?} and \emph{If discrepancies arise, which factors have the most significant impact?}

To address these questions, we conducted user studies using the images from ExDark dataset~\cite{LOH201930}, comparing dim images with outputs from 14 low-light enhancement methods. These methods include one optimization-based approach  (LIME~\citep{DBLP:journals/tip/GuoLL17}), 7 supervised methods (Bread~\citep{DBLP:journals/ijcv/GuoH23}, Kind~\citep{DBLP:conf/mm/ZhangZG19}, Retinexformer~\citep{cai2023retinexformer}, IAT~\citep{cui2022need}, SNR~\citep{xu2022snr}, LLFlow~\citep{DBLP:conf/aaai/WangWYLCK22} and PyDiff~\citep{DBLP:conf/ijcai/ZhouYY23}), and 6 unsupervised methods (QuadPrior~\citep{quadprior}, LightenDiffusion~\citep{Jiang_2024_ECCV}, SCI~\citep{ma2022toward}, ZeroDCE~\citep{Guo_2020_CVPR} PairLIE~\citep{fu2023learning} and NeRCo~\citep{DBLP:conf/iccv/YangDWLZ23} ). 
Inspired by Chatbot Arena~\citep{DBLP:conf/icml/ChiangZ0ALLZ0JG24}, we presented users with pairs of images generated by different enhancement methods, and randomly selected one of five aspects—overall quality, illumination, noise/artifacts, blurriness, or color. Participants were then asked to choose the better option based on the selected criterion. To quantify user preferences across all methods, we employed the Elo rating system to convert these pairwise comparisons into a comprehensive rating. Further details of the study can be found in the appendix.
 
As depicted in  Fig~\ref{fig:Elo} (a), the direct evaluation of detection performance shows a strong correlation with user preferences from the study. The Spearman correlation coefficient $r$ is 0.703, indicating a robust positive relationship that suggests a general consistency between detection scores and user-assigned image ratings across different methods. This correlation is statistically significant, with a p-value of 0.0035. However, several outliers are evident in the figure: 1) While IAT demonstrates strong detection performance, it ranks 13th in overall user preference. As shown in Fig.~\ref{fig:Elo}(b), IAT's color rendering is particularly unappealing, a crucial factor to human perception that may be overlooked by detectors trained with extensive color jitter augmentation. 2) The input image ranks 8th in the Elo rating, but achieves third-best detection performance. As illustrated in Fig.~\ref{fig:Elo}(b), although the dim image is poorly-illuminated, its noise and JPEG artifacts are (of course) less noticeable in the darker areas. In contrast, some enhancement methods (\emph{e.g.}, NeRCo) may inadvertently introduce additional artifacts during the enhancement process, resulting in lower detection performance despite potential improvement in illumination.

\begin{figure}[t]
    \centering
    \includegraphics[width=0.315\linewidth]{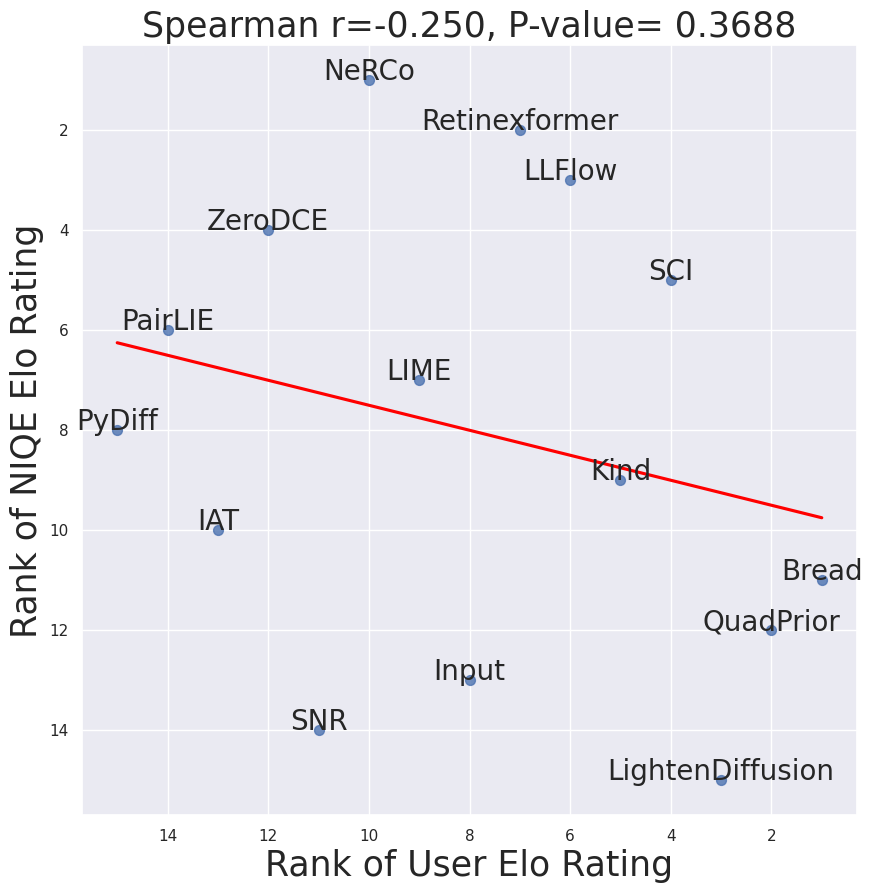}
    \includegraphics[width=0.315\linewidth]{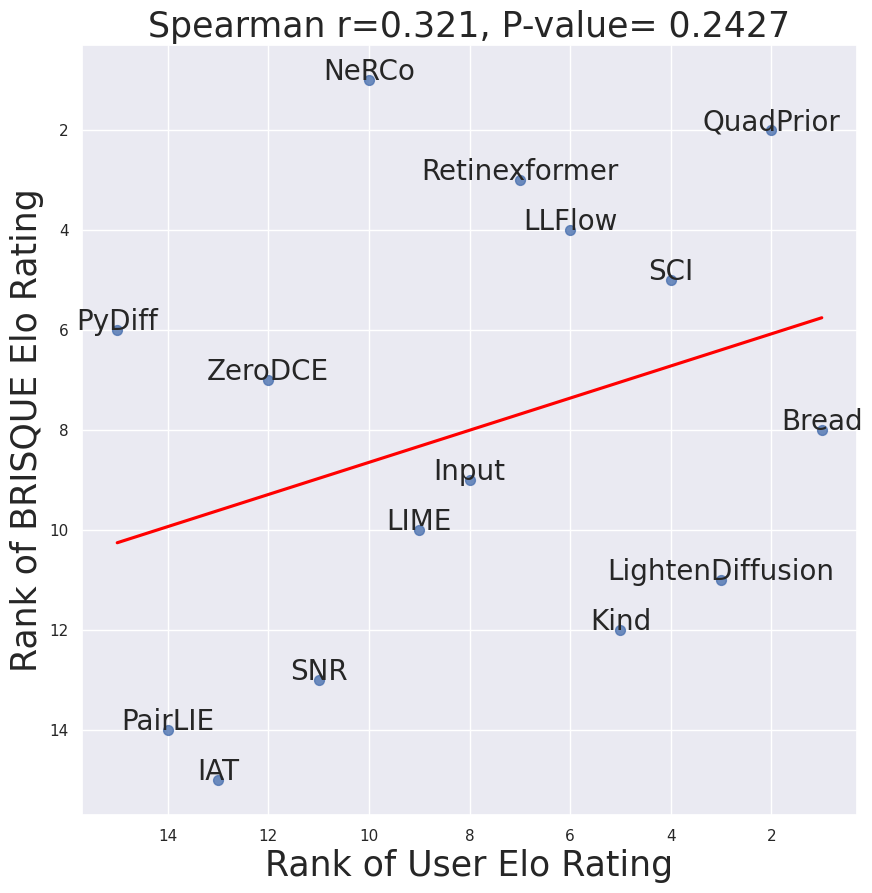}
    \includegraphics[width=0.315\linewidth]{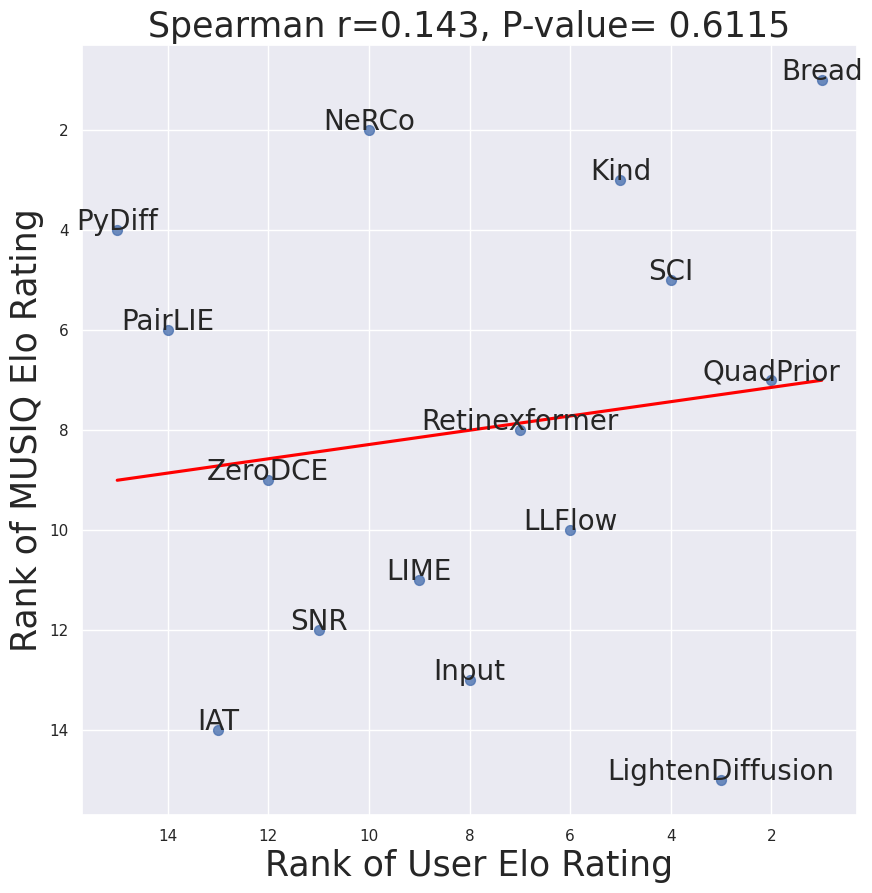}

    \begin{subfigure}{0.315\linewidth}
        \centering
        \subcaption{NIQE}
    \end{subfigure}
    \begin{subfigure}{0.315\linewidth}
        \centering
        \subcaption{BRISQUE}
    \end{subfigure}
    \begin{subfigure}{0.315\linewidth}
        \centering
        \subcaption{MUSIQ}
    \end{subfigure}
    
    \includegraphics[width=0.315\linewidth]{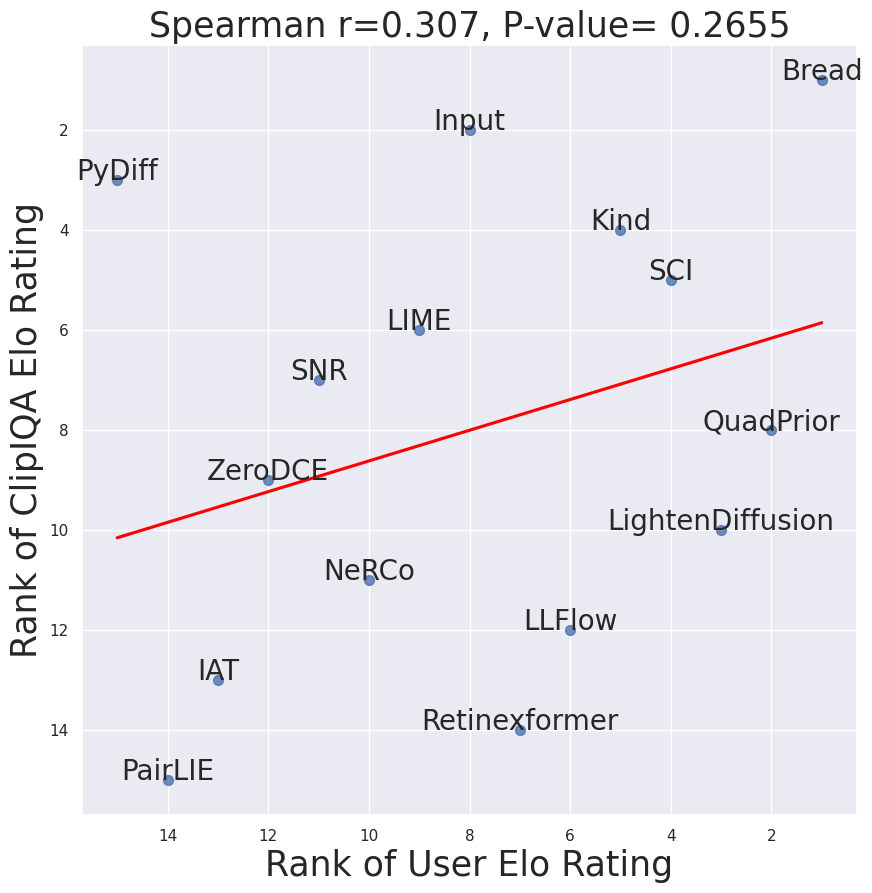}
    \includegraphics[width=0.315\linewidth]{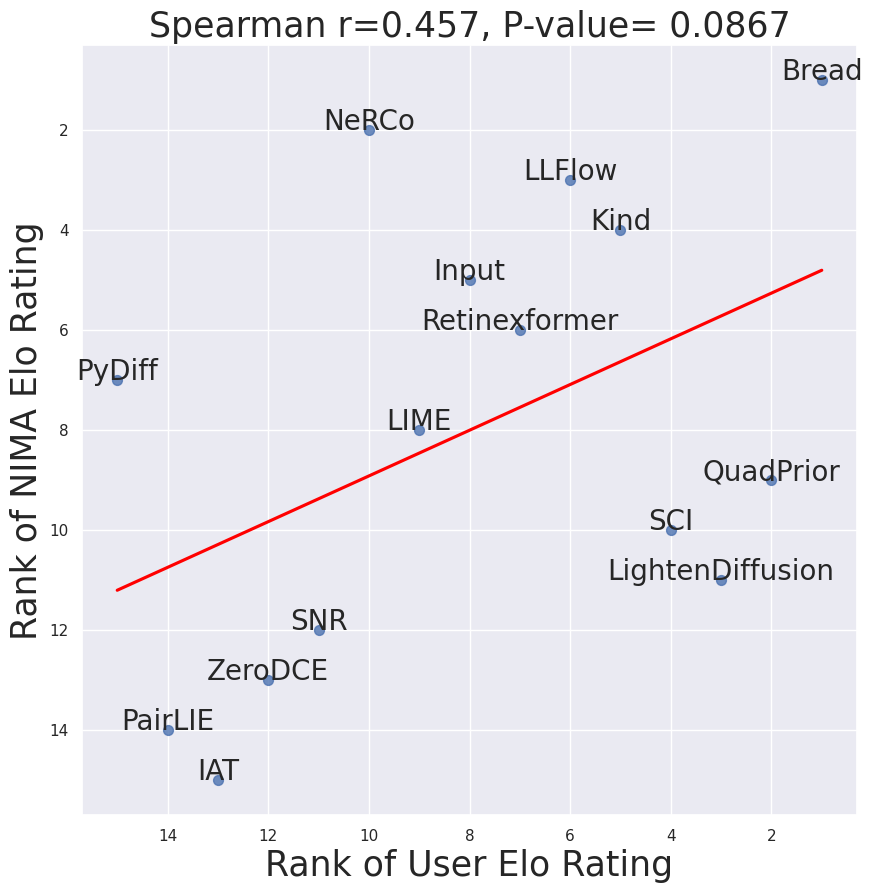}
    \includegraphics[width=0.315\linewidth]{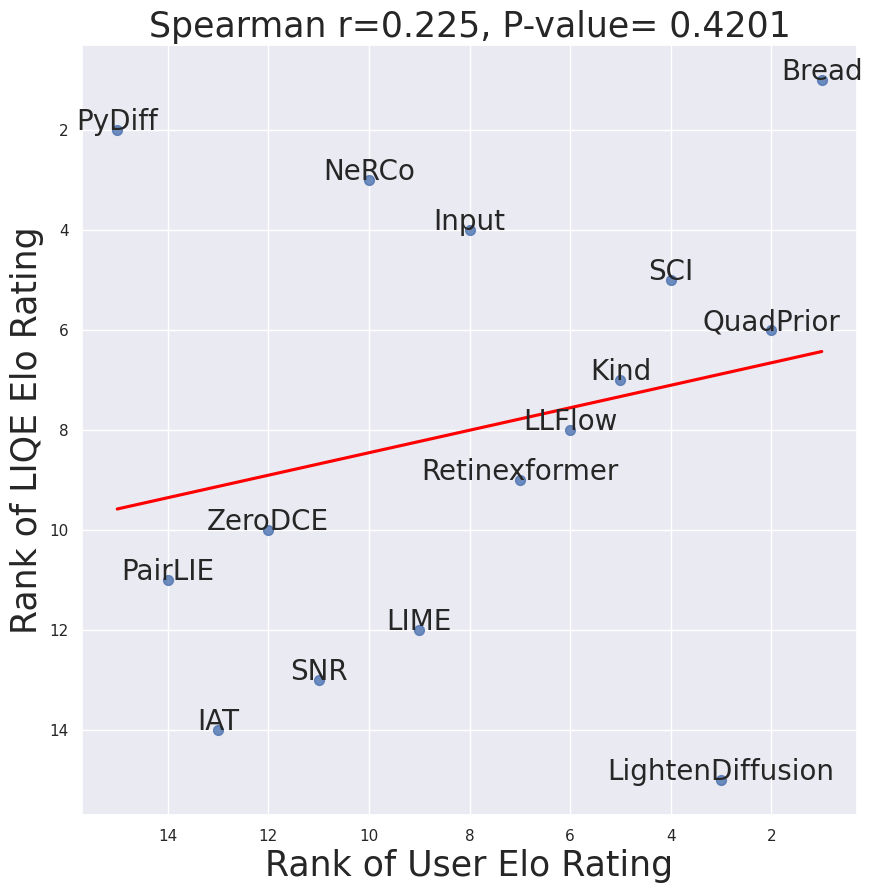}

    \begin{subfigure}{0.315\linewidth}
        \centering
        \subcaption{ClipIQA}
    \end{subfigure}
    \begin{subfigure}{0.315\linewidth}
        \centering
        \subcaption{NIMA}
    \end{subfigure}
    \begin{subfigure}{0.315\linewidth}
        \centering
        \subcaption{LIQE}
    \end{subfigure}
    \caption{Correlations between user preference and popular IQA approaches.}
    \label{fig:iqa_corr}
    \vspace{-15pt}
\end{figure}

Consequently, direct evaluation with detectors is less sensitive to color shift and poor illumination, but is more sensitive to noise and artifact. Despite these discrepancies, the overall correlation indicates that direct detection performance can serve as a reasonable proxy for assessing enhancement quality. To illustrate the advantage of using direct evaluation scheme, we selected 6 popular IQA/IAA methods in low-light enhancement, including NIQE~\citep{Mittal2013Making}, BRISQUE~\citep{DBLP:journals/tip/MittalMB12}, MUSIQ~\citep{DBLP:conf/cvpr/SahaMB23}, ClipIQA~\citep{wang2022exploring}, NIMA~\citep{DBLP:journals/tip/TalebiM18} and LIQE~\citep{zhang2023liqe}. We fed the same input images used in the user study for benchmarking. The results can be found in Fig.~\ref{fig:iqa_corr}. BRISQUE, ClipIQA, LIQE, and MUSIQ tend to favor the outputs from PyDiff and NeRCo while overlooking the performance of LightenDiffusion. Among these methods, NIMA reaches the best correlation with human preferences, with a Spearman $r$ of 0.457. However, the alignment between these quality assessment approaches and human perception remains inferior to that of the detection-based evaluation, highlighting the reliability of direct detection performance as an 
assessment metric. 

\begin{figure}[t]
    \centering
    \includegraphics[width=0.35\linewidth]{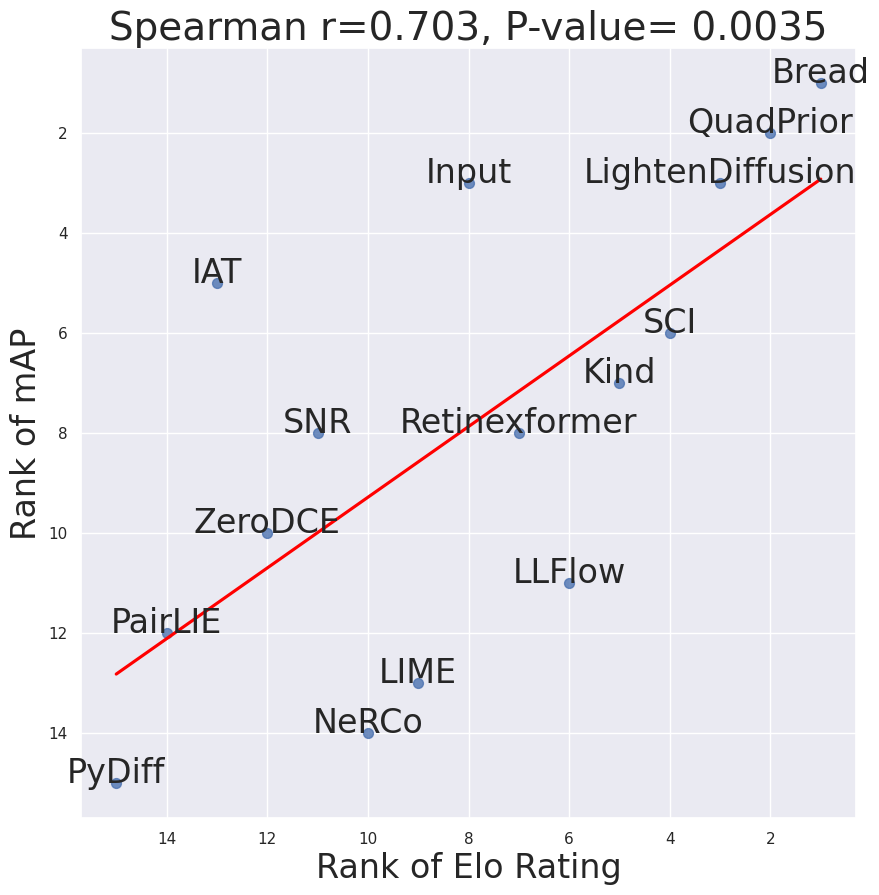}
    \includegraphics[width=0.58\linewidth]{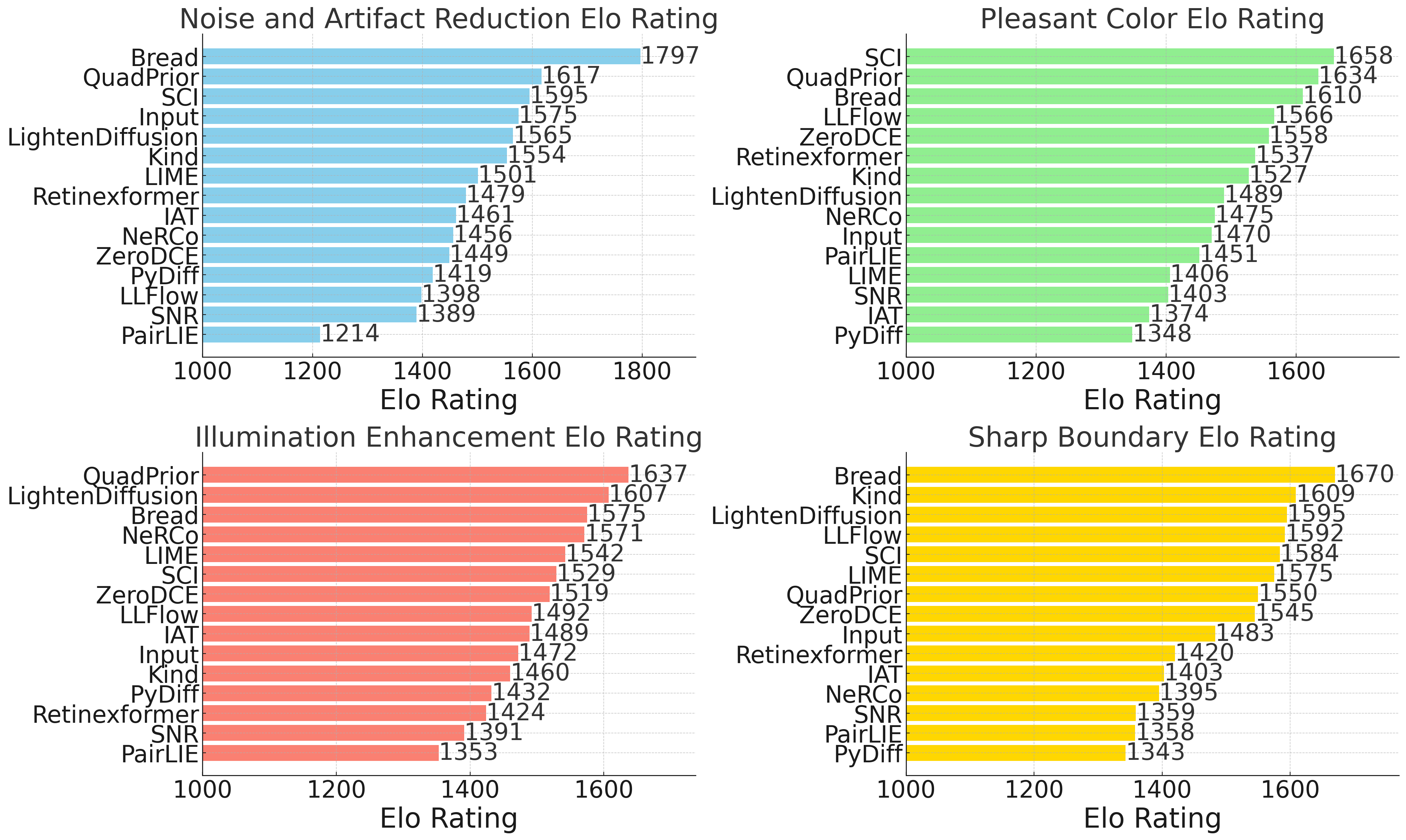}

    \begin{subfigure}{0.40\linewidth}
        \centering
        \subcaption{}
    \end{subfigure}
    \begin{subfigure}{0.58\linewidth}
        \centering
        \subcaption{}
    \end{subfigure}
    
    \caption{User preference study. (a) plots the rank of the overall user preference (Elo Rating) in relation to detection performance (mAP). (b) depicts the Elo Ratings respective for noise/artifact reduction, illumination enhancement, color restoration, and boundary sharpness. }
    \label{fig:Elo}
\end{figure}

\section{Towards Label-free Evaluation through LIME-eval}

Previous experimental validation still relies on annotated datasets. The dependency on annotated labels presents a series of significant hurdles. Primarily, securing precise annotations for low-light images is notably more difficult and time-consuming than for well-lit images due to the inherent challenges associated with low visibility. The reduced contrast and clarity in low-light conditions often lead to unclear object boundaries and categories, elevating the potential for inaccuracies in annotations. Furthermore, the endeavor to annotate an large-scale dataset of low-light images for establishing a dependable benchmark demands considerable labor and expense, thereby constraining the scalability and practicality of such evaluative methods.

Given the reliance on annotated datasets in these experiments, there emerges a pressing need for an evaluation methodology that operates independently of labels. A label-independent evaluation approach would streamline the assessment of low-light image enhancement techniques and broaden the applicability across diverse and unlabeled datasets. Consequently, exploring and developing an evaluation strategy that transcends the need for annotated datasets becomes a critical next step in advancing the field of low-light image enhancement. In what follows, we shall introduce our LIME-Eval, a label-free evaluation metric, as a pioneering exploration of this problem. 

\subsection{Detection-oriented Energy-based Modeling}
\label{sec:labeling}
The energy-based model (EBM) ~\citep{lecun2006tutorial} can map data point $x$ with any dimension into a scalar through an energy function $Z(x): \mathbb{R}^D \rightarrow \mathbb{R}$. To transfer the energy function into a probability density function $p(x)$, one could adapt the Gibbs distribution as follows:
\begin{equation}
    p(y\mid x)=\frac{e^{-Z(x,y)/T}}{\int_{y^{\prime}}e^{-Z(x,y^{\prime})/T}}=\frac{e^{-Z(x,y)/T}}{e^{-Z(x)/T}},
\end{equation}
where $\int_{y^{\prime}}e^{-Z(x,y^{\prime})/T}$ is the partition function by marginalizing over label $y$, and $T$ is a positive temperature constant. Now the Gibbs free energy $Z(x)$ at the data point $x$ with the negative of the log partition function can be  written as:
\begin{equation}
    Z(x)=-T\cdot\log\int_{y^{\prime}}e^{-Z(x,y^{\prime})/T}.
\end{equation}
The inherent connection between energy-based models and discriminative models has been explored in \citep{DBLP:conf/iclr/GrathwohlWJD0S20,peng2024energy}. Consider a $K$-category classifier $f$, which maps input vector $x$ into $K$ logits, with the softmax function, we can parameterize a categorical distribution via:
\begin{equation}
    p(y\mid x)=\frac{e^{f_y(x)/T}}{\sum_{k=1}^Ke^{f_{k}(x)/T}},
\end{equation}
where $f_{y}(\cdot)$ denotes logit corresponding to $y$-th term of $f(x)$. Thus, the energy function can be expressed as:
\begin{equation}
  Z(x)=T\cdot\log\sum_{j=k}^Ke^{f_{k}(x)/T}.
  \label{eq:final}
\end{equation}

\begin{figure}[t]
    \centering
    \includegraphics[width=\linewidth]{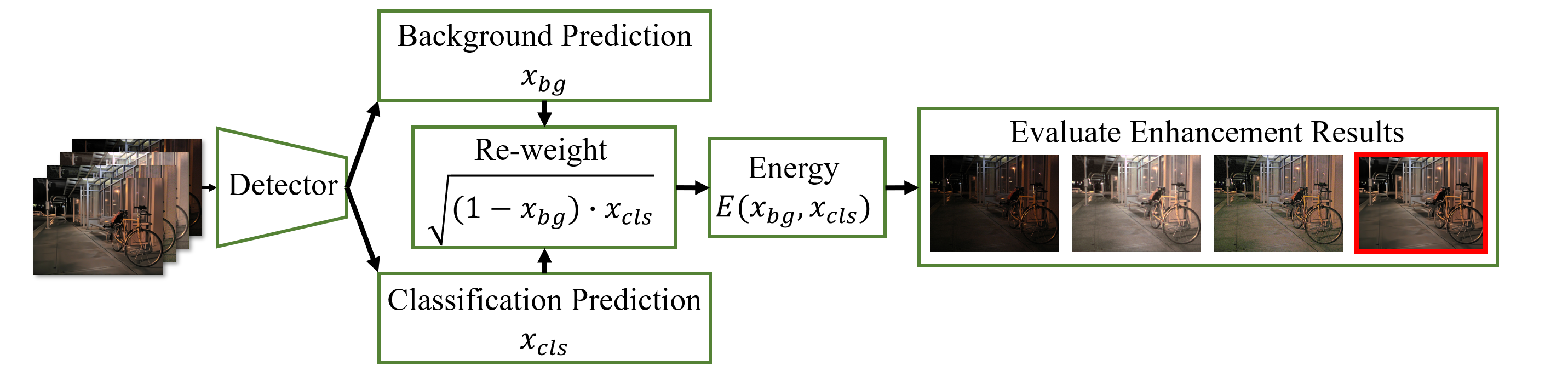}
    \caption{The pipeline of our proposed LIME-Eval.}
    \label{fig:pipeline}
\vspace{-15pt}
\end{figure}

The above modeling can only be applied to the task of classification, which has been adapted to the AutoEval~\citep{peng2024energy}. The classification task only involved one overall prediction. To adopt it into our target task, \textit{i.e.} object detection\footnote{For simplicity, here we omit the multi-scale outputs and consider the output as heatmaps in $H\times W$.}, which contains both classification output $x_{cls} \in \mathbb{R}^{K\times H\times W}$ and object output $x_{bg} \in \mathbb{R}^{H \times W}$, we propose to fuse the logits in $x_{cls}$ with the objectness information $x_{bg}$ into re-weighted energy $x_r$ as follows:
\begin{equation}
    \label{eq:reweight}
    x_{r}^{y} = \sqrt{x_{cls}^{y} \cdot (1 - x_{bg})},
\end{equation}
where $x_{r}^{y}$ denotes the $y$-th logit of $x_r$. After that, we integrate Eq.~(\ref{eq:final}) into final evaluation function $E(x_{cls}, x_{bg})$ as follows:
\begin{equation}
    \label{eq:evaluation_function}
    E(x_{cls}, x_{bg}) = -\sum_{i,j}^{H, W}{T \log(\sum_{y} e^{x_r^{y,i,j} / T} )},
\end{equation}
where $x_{r}^{y, i, j}$ denotes logit corresponding to $y$-th term of $x_{r}$ at $(i,j)$. This indicator transforms spatial confidence information into a distribution measure, which can be further aggregated over the dataset for a dataset-level metric. The overall pipeline of our LIME-Eval is illustrated in Fig.~\ref{fig:pipeline}. After extracting background prediction $x_{bg}$ and classification prediction $x_{cls}$, the two feature maps are fused via $\sqrt{x_{cls}(1-x_{bg})}$. The energy is calculated as in Eq.~(\ref{eq:evaluation_function}). Finally, we identify the image with the lowest energy value as the optimal one.

\subsection{Experimental Validation}

\textbf{Correlation Studies on Synthesised Datasets.} Having the evaluation function defined, it becomes feasible to assess images without relying on labels. To show that our proposed energy metric aligns with detection performance, we synthesized images from the validation set of MS-COCO with low-light-related distortion.  
\begin{figure}[t]
    \centering
    \includegraphics[width=\linewidth]{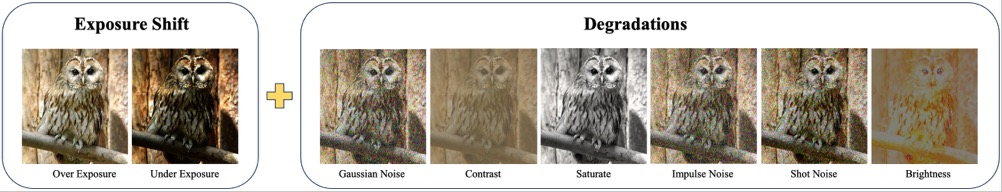}
    \caption{A visualization of synthetic setting. More details can be found in Appendix.}
    \label{fig:syn}
\vspace{-15pt}
\end{figure}
To obtain a more accurate approximation of the mean Average Precision (mAP), we begin by pre-calibrating the images using a synthetic dataset derived from the validation split of the MS-COCO dataset. MS-COCO, a large-scale dataset for object detection, encompasses 80 object categories, providing a robust foundation for evaluating detection performance. The validation set of MS-COCO consists of 5,000 images, capturing a wide range of everyday scenes. 


Inspired by typical low-light enhancers~\citep{DBLP:journals/ijcv/GuoH23, DBLP:journals/tip/GuoLL17}, which first perform exposure correction and then handle degradation, our synthesis pipeline is similarly divided into two primary phases to address two types of distortions: 1) \emph{Exposure Shift}. The first phase focuses on the prevalent issues of overexposure and underexposure commonly observed in low-light enhancement outputs. We employ gamma correction to simulate these exposure shifts; and 2) \emph{Degradation}. The second phase replicates degradations such as ineffective noise suppression, leading to either noise persistence or excessive image smoothing, as well as color distortions. To simulate these effects, we introduce impulse, shot, and Gaussian noises for the former, and employ Gaussian blur for the latter. Further, we adopt strategies to reduce saturation and brightness, mimicking color distortions.

\begin{figure}[t]
    \centering
    \includegraphics[width=.24\linewidth]{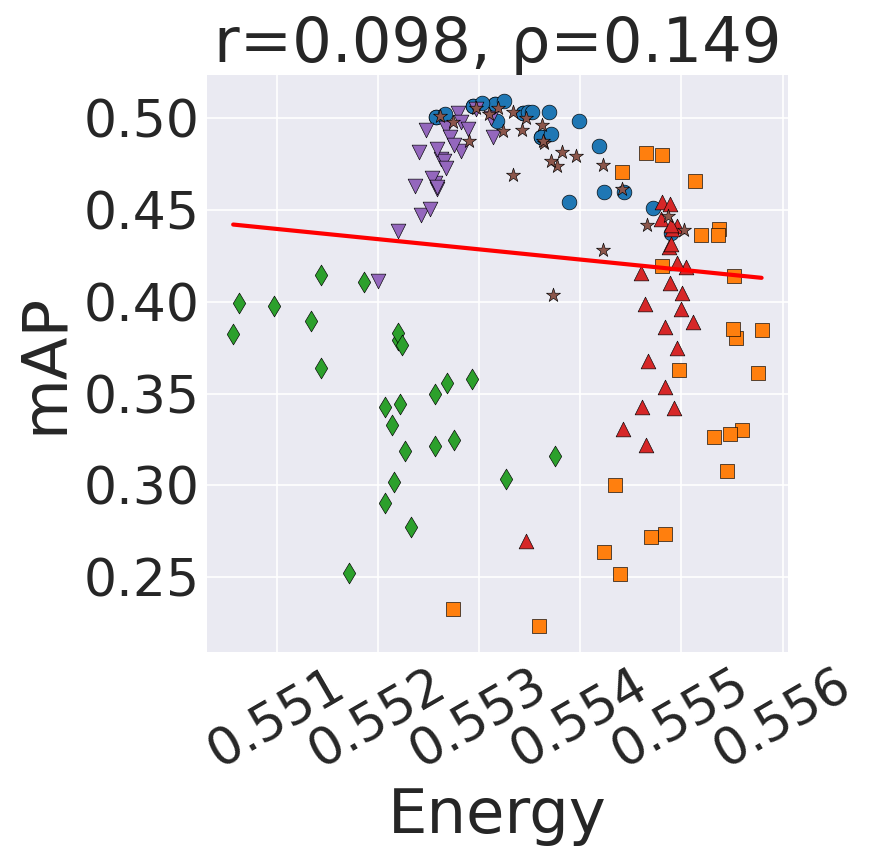}
    \includegraphics[width=.23\linewidth]{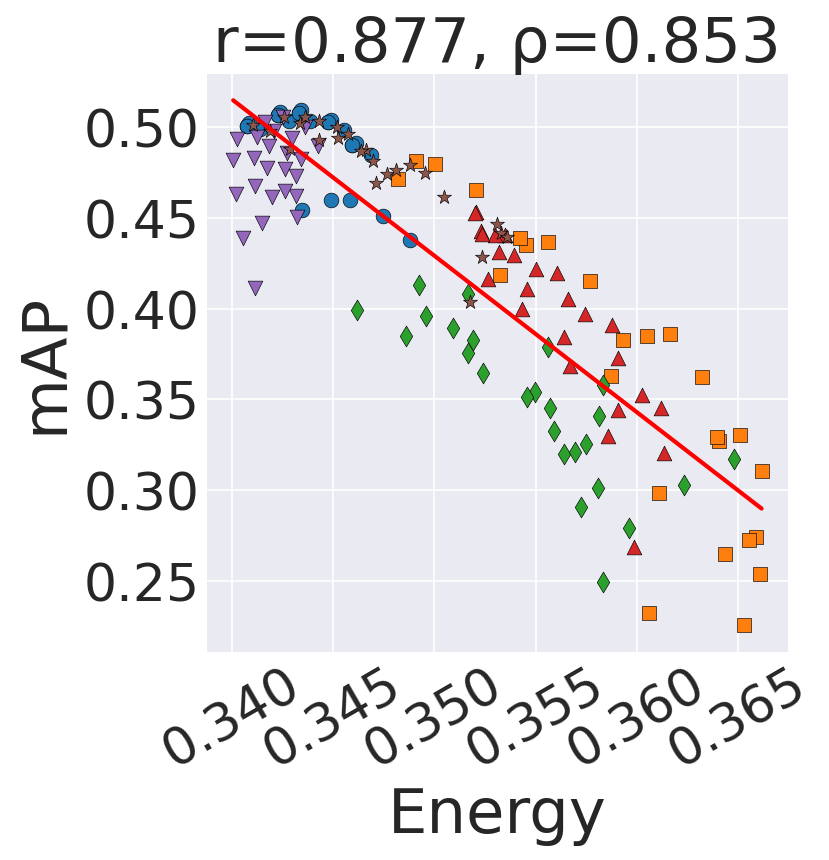}
    \includegraphics[width=.23\linewidth]{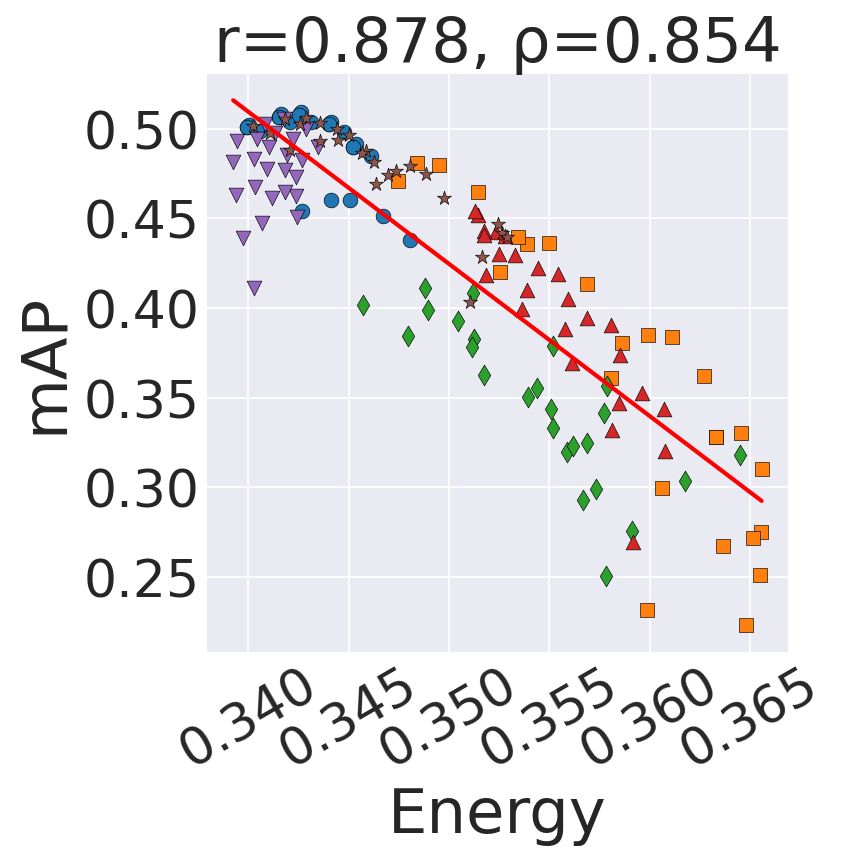}
    \includegraphics[width=.23\linewidth]{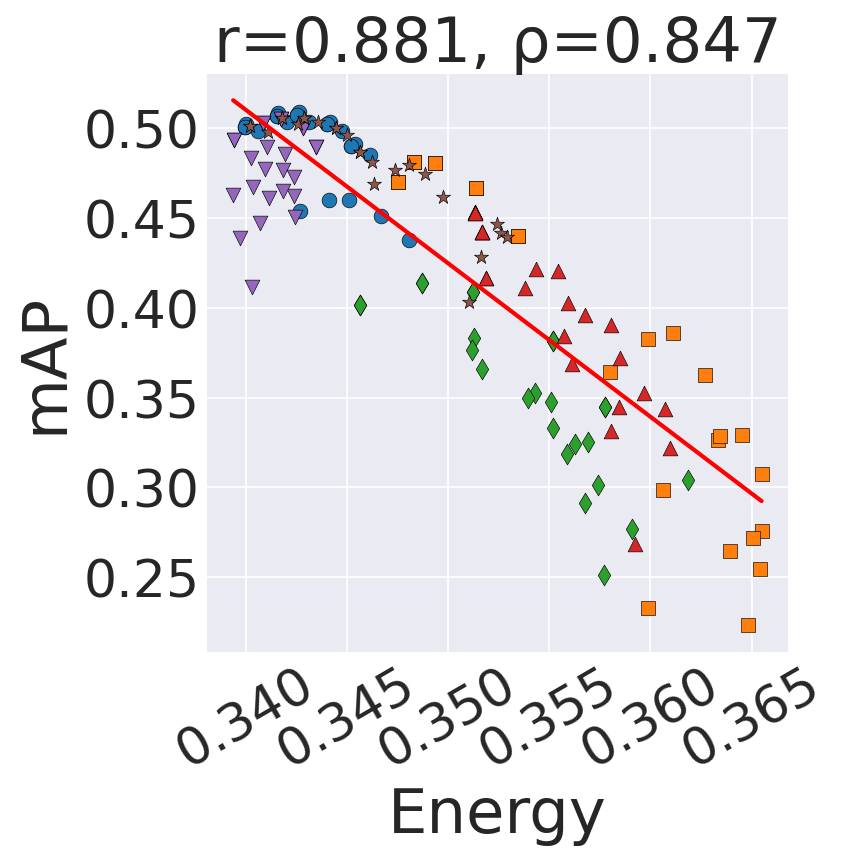}

    \begin{subfigure}{0.24\linewidth}
        \centering
        \subcaption{$T = 0.1$}
    \end{subfigure}
    \begin{subfigure}{0.24\linewidth}
        \centering
         \subcaption{$T = 0.01$}
    \end{subfigure}
    \begin{subfigure}{0.24\linewidth}
        \centering
       \subcaption{$T = 0.001$}
    \end{subfigure}
    \begin{subfigure}{0.24\linewidth}
        \centering
        \subcaption{$T = 0.0001$}
    \end{subfigure}
    \vspace{-5pt}
    \caption{Energy versus mAP under different temperatures on our synthesized dataset. Data points as \textcolor{Type1Blue}{\large$\circ$}, \textcolor{Type2Orange}{$\square$}, \textcolor{Type3Green}{\large$\diamond$}, \textcolor{Type4Red}{$\triangle$},\textcolor{Type5Purple}{$\triangledown$},\textcolor{Type6Brown}{\large$\star$} refers to over-smooth, Gaussian Noise, impulse noise, shot noise, brightness adjustment and saturate adjustment. The calibrated energy function is plotted in red line. }
    \vspace{-15pt}
    \label{fig:regression_temperture}
\end{figure}

A detailed depiction of the synthesizing process is illustrated in Fig.~\ref{fig:syn}. To quantitatively evaluate the performance of our energy metric, we employed two statistical measures: Pearson's correlation coefficient ($\rho$)  and Spearman's rank correlation coefficient ($r$). Pearson's correlation ($\rho$) measures the linear relationship between the energy metric values and detection performance scores, providing insight into how well the metric predicts actual performance improvements. Spearman's rank correlation ($r$), on the other hand, assesses how well the relationship between the energy metric and detection performance follows a monotonic function. This is particularly useful for understanding the metric's ability to rank enhancement methods accroding to their impact on detection performance, irrespective of the linearity of the relationship. A YOLOX-x model trained on the MS-COCO dataset is adopted as $f$ (aforementioned classifier). As can be seen from Fig. \ref{fig:regression_temperture}, our proposed method shows a strong correlation with mAP ($r = 0.881$, $\rho = 0.847$), indicating that the proposed evaluation function aligns closely with actual mAP, even without the help of labels. 

\textbf{Consensus with Human-preference.} We also employ user preference from LIME-Bench to benchmark the performance of the proposed method. As reported in Tab.~\ref{tab:hf}, our method exhibits a stronger correlation with human preferences compared to previous image quality assessment methods, demonstrating its superior alignment with perceptual quality judgments.

\begin{table}[t]
    \centering
    \caption{Spearman correlation comparison between IQA methods and our LIME-Eval. The best is in \textbf{bold} and the second-best is \underline{underlined}.}
    \begin{tabular}{c|c|c|c|c|c|c|c}
    \hline
      Method & BRISQUE & NIQE & NIMA & LIQE & MUSIQ & ClipIQA &LIME-Eval \\
        \hline
     Spearman $r$ &  0.321 & -0.250 & \underline{0.457} & 0.225 & 0.143 & 0.307 & \textbf{0.593} \\
        \hline
    \end{tabular}
    \vspace{-15pt}
    \label{tab:hf}
\end{table}

\begin{table}[t]
    \centering
    \caption{Quantitative results on the LOL-v2 real dataset. We retrained Retinexformer based on the code released by the authors with their recommended training recipe and dataset split.}
    \begin{tabular}{c|c|c|c}
    \hline
        &LPIPS$\downarrow $& mAP$\uparrow $& AP50$\uparrow $\\
        \hline
        RetinexFormer & 0.1863  & 34.0 & 64.2\\
        \rowcolor{Gray}
        RetinexFormer + LIME-Eval & 0.1625\textbf{$(\mathbf{0.0238\downarrow})$} & 34.7 {$\mathbf{(0.7\uparrow})$} & 64.9 \textbf{$(\mathbf{0.7\uparrow})$}\\
        \hline
    \end{tabular}
    \vspace{-15pt}
    \label{tab:energy_experiment}
\end{table}

\begin{figure}[t]
    \centering

    \includegraphics[width=\linewidth]{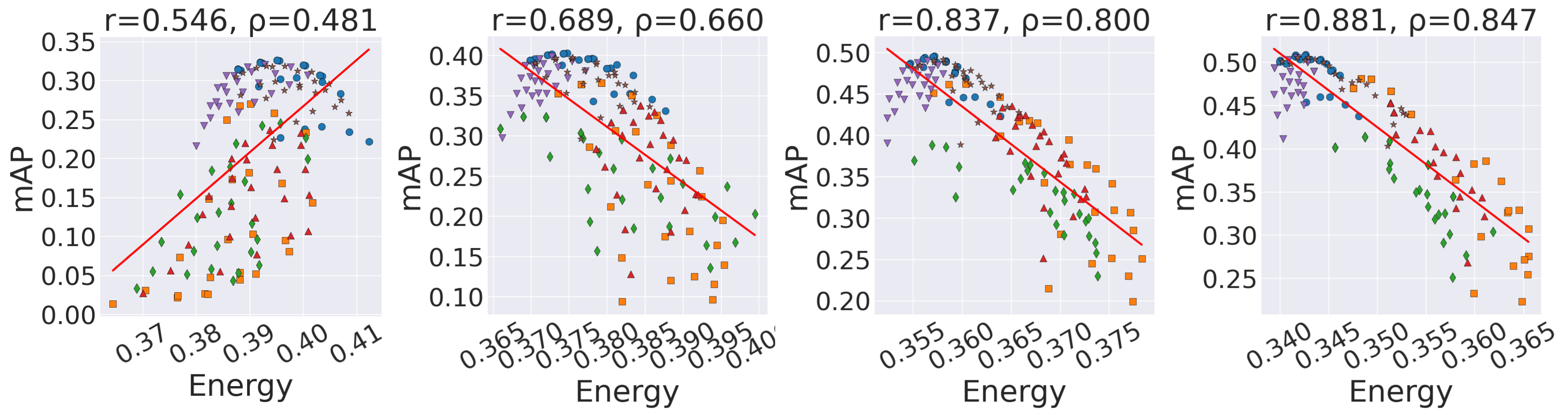}

    \begin{subfigure}{0.24\linewidth}
        \centering
        \subcaption{YOLOX-Tiny}
    \end{subfigure}
    \begin{subfigure}{0.24\linewidth}
        \centering
         \subcaption{YOLOX-S}
    \end{subfigure}
    \begin{subfigure}{0.24\linewidth}
        \centering
       \subcaption{YOLOX-L}
    \end{subfigure}
    \begin{subfigure}{0.24\linewidth}
        \centering
        \subcaption{YOLOX-X}
    \end{subfigure}
	\vspace{-5pt}
    \caption{Energy versus mAP under different base detection on our synthesized dataset. Data points as \textcolor{Type1Blue}{\large$\circ$}, \textcolor{Type2Orange}{$\square$}, \textcolor{Type3Green}{\large$\diamond$}, \textcolor{Type4Red}{$\triangle$},\textcolor{Type5Purple}{$\triangledown$},\textcolor{Type6Brown}{\large$\star$} refers to over-smooth, Gaussian Noise, impulse noise, shot noise, brightness adjustment and saturate adjustment.  The calibrated energy function is plotted in red line. The result with YOLOX-x performs best among choices.
    }
    \vspace{-15pt}
    \label{fig:regression_model}
\end{figure}

\subsection{Backpropgation of Energy Helps Enhancers}
\label{sec:exp_bp}
 
Since our energy function is differentiable, it can serve as a loss function to provide additional regularization for low-light enhancers. We demonstrate this by integrating our energy function into the training process of Retinexformer.
 As shown in Tab.~\ref{tab:energy_experiment}, the model enhanced with our energy function enjoys a favorable gains in both low-level metric(LPIPS) and downstream detection metrics (mAP and AP50). These results indicate that the energy function contributes effectively to regularization, aiding the enhancement process.  Qualitative comparison can be found in the appendix.

\subsubsection{Ablation Study}

\textbf{The Effect of Hyper-parameter $T$.} Given that our framework relies on a single hyper-parameter, $T$, we have conducted a series of experiments to assess its sensitivity and impact on performance. The experimental results are systematically presented in Fig.~\ref{fig:regression_temperture}. The findings show that our energy-based metric maintains a strong correlation across a range of values from 0.01 to 0.0001, indicating that the method stays stable over a broad spectrum of temperature settings.  

\textbf{The Effect of Model Size.} We also explored the impact of model size on the performance of our framework by experimenting with different versions of the YOLOX architecture: YOLOX-Tiny, YOLOX-s, YOLOX-l, and YOLOX-x. This investigation aims to understand how the size of the base detector influences detection accuracy, processing speed, and overall system efficiency within our enhanced low-light image evaluation setup. The outcomes of these experiments, which detail the trade-offs associated with each model size, are documented in Fig.~\ref{fig:regression_model}. As we can observe from the figure, the larger the model, the stronger the correlation energy with the mAP will be. When we scale the model back to YOLOX-Tiny, the connection between energy and mAP vanishes.   

\section{Conclusion}
In this study, we have provided a comprehensive evaluation of low-light image enhancement techniques, with a particular focus on the application of an energy-based model for assessment. Our investigation, rooted in the analysis of object detection on the ExDark dataset and utilizing the medium version of YOLOX as the base detector, has illuminated the limitations inherent in traditional evaluation methodologies that rely on retraining recognition models on enhanced images. We have demonstrated that such approaches can lead to overfitting, thus skewing the fairness and accuracy of evaluations. 

By adopting an energy-based model for evaluation, we have introduced a novel framework that sidesteps the pitfalls of overfitting and offers a more equitable measure of an enhancement technique's effectiveness. Our findings reveal that this method not only provides a more accurate reflection of an algorithm's performance in enhancing low-light images but also presents a promising avenue for future research in image processing and evaluation. The implications of our study extend beyond the immediate scope of low-light image enhancement, suggesting a broader applicability of energy-based models in the evaluation of image processing techniques.

\bibliography{egbib}
\bibliographystyle{iclr2025_conference}
\clearpage

\appendix
\section{Appendix}
\subsection{LIME-bench Details}
Our LIME-bench collects data through an online user survey. A screenshot of our system can be found in Fig. Specifically, we randomly select an image, a particular attribute, and two enhancement methods (including the original input). Users are then asked to choose between four options: Image 1 is better, Image 2 is better, both are good, or both are bad. We adopt Elo rating system to obtain final rating for each method. For a pair of user preference, if user can tell which one is better, then we update the score with $k$ set to 16. However, when user voted for "both are good/well", we treat it as the two competitor both win/lose from the original input, since this requires 2 times of score update, we down-weighted the $k$ to 8 in this situation.
\begin{figure}[t]
    \centering
    \includegraphics[width=0.9\linewidth]{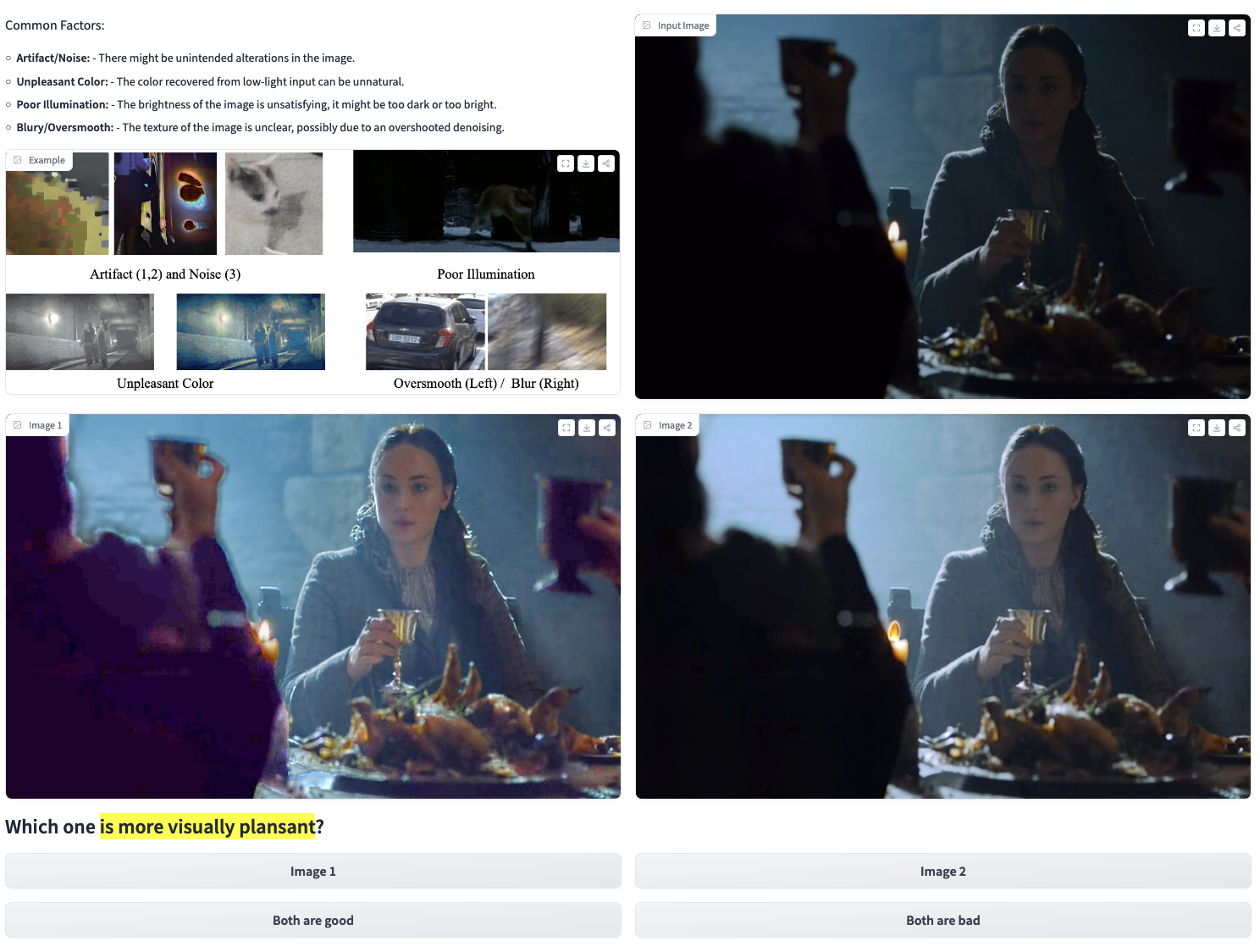}
    \caption{A screenshot of our online user survey system.}
    \label{fig:web}
\end{figure}
\subsection{Implementation Details}
In this work, we use PyTorch to implement our LIME-eval framework. All of our experiments are carried out on NVIDIA RTX3090 GPUs. The detectors and enhancers are trained with the code and configuration (optimizer, learning rate random seeds, etc.) provided by the authors to provide a best-effort fair comparison, except for Tab.~\ref{tab:boost_with_aug}, where we carefully tuned the parameters for the best performance since no existing training recipe for us to follow.

\subsubsection{Data Synthesis}
The data synthesis pipeline we have used comprises two types of distortions, the settings of which are as follows:
\begin{enumerate}
    \item Exposure Shifts
    \begin{itemize}
        \item Under Exposure. Gamma correction with $\gamma=1.5, 2$
        \item Over Exposure. Gamma correction with $\gamma=0.75, 0.5$
        \item Original Exposure. (Gamma correction with $\gamma=1$)
    \end{itemize}
    \item Degradation
    \begin{itemize}
        \item Gaussian Blur with $\sigma_s = 0.1, 0.2, 0.4, 0.8, 1.6$ 
        \item Gaussian Noise under level $5, 10, 15, 20, 25$
        \item Impulse Noise, amount $= 0.01, 0.025, 0.05, 0.075, 0.1$
        \item Shot Noise under level $60, 45, 30, 20, 12$
        \item Brightness distortion. First, convert the image into HSV color space, then add $0.1, 0.2, 0.3, 0.4, 0.5$ to $V$.
        \item Saturate distortion. First, convert the image into HSV color space, then scale component $S$ with $\alpha S + \beta$, where $\alpha = 0.3, 0.1, 2, 5, 20$ and $\beta = 0, 0, 0, 0.1, 0.2$ 
    \end{itemize}
\end{enumerate}

For every image, we first select a degradation and then perform an exposure shift. In this way, we generate 150 distorted datasets for correlation analysis.

\subsection{More Qualitative Comparisons}
\begin{figure}[!b]
    \centering
    
    \includegraphics[width=.19\linewidth]{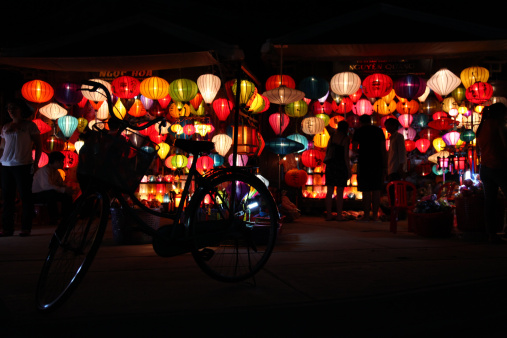}
    \includegraphics[width=.19\linewidth]{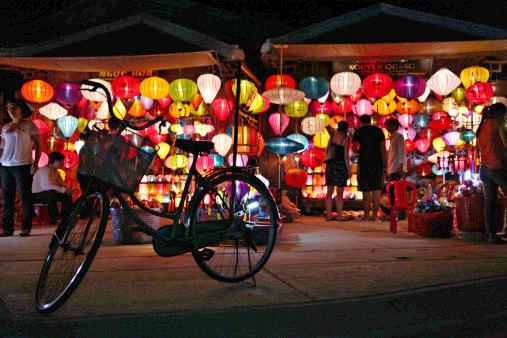}
    \includegraphics[width=.19\linewidth]{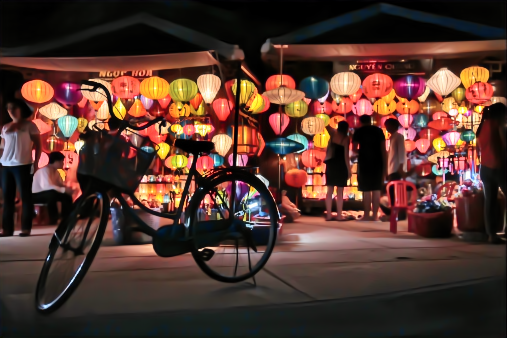}
    \includegraphics[width=.19\linewidth]{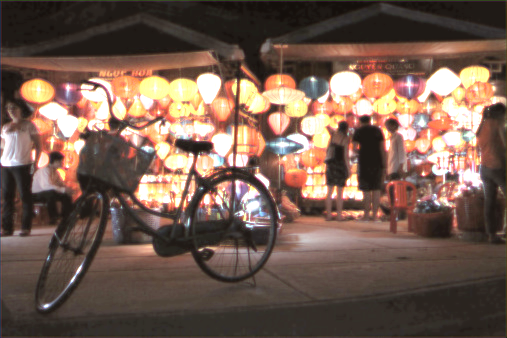}
    \includegraphics[width=.19\linewidth]{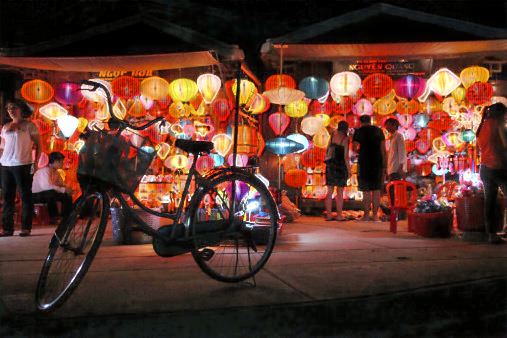}

    \includegraphics[width=.19\linewidth]{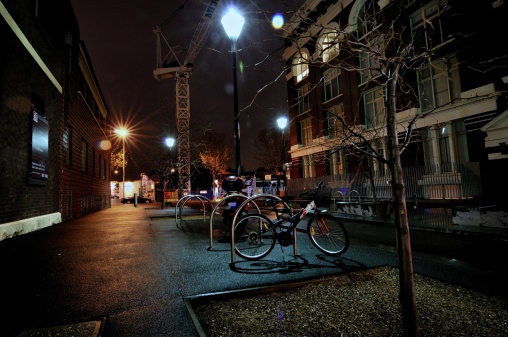}
    \includegraphics[width=.19\linewidth]{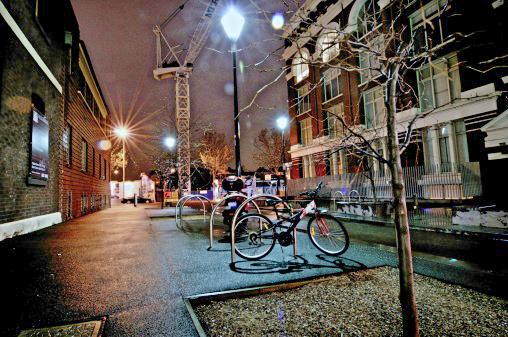}
    \includegraphics[width=.19\linewidth]{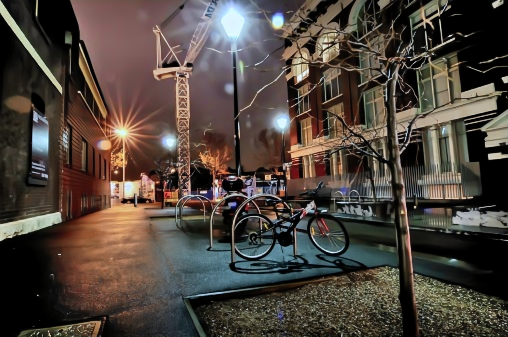}
    \includegraphics[width=.19\linewidth]{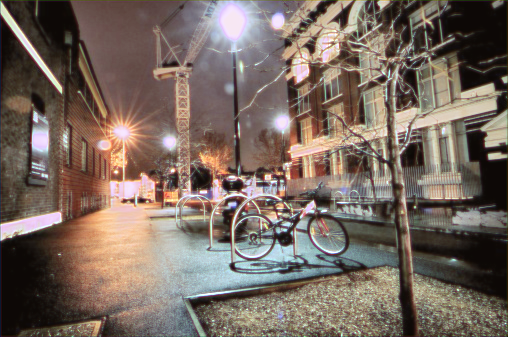}
    \includegraphics[width=.19\linewidth]{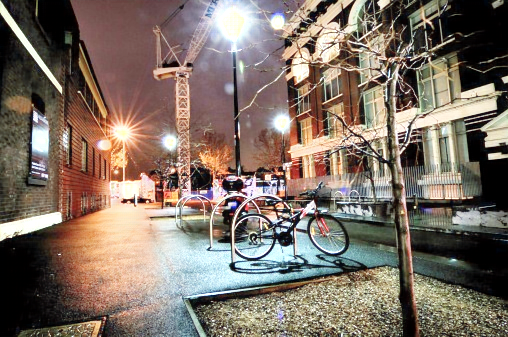}

    \includegraphics[width=.19\linewidth]{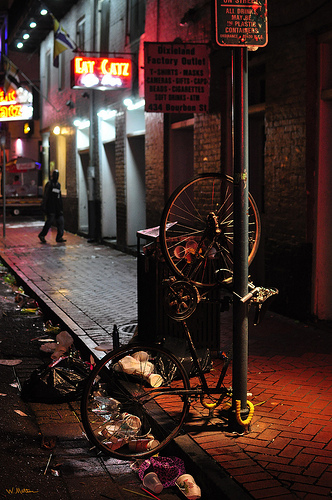}
    \includegraphics[width=.19\linewidth]{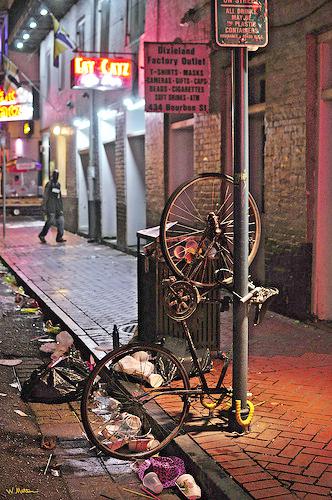}
    \includegraphics[width=.19\linewidth]{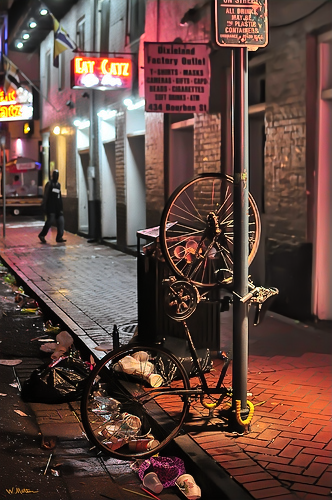}
    \includegraphics[width=.19\linewidth]{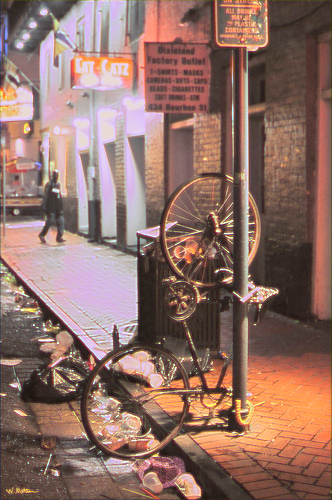}
    \includegraphics[width=.19\linewidth]{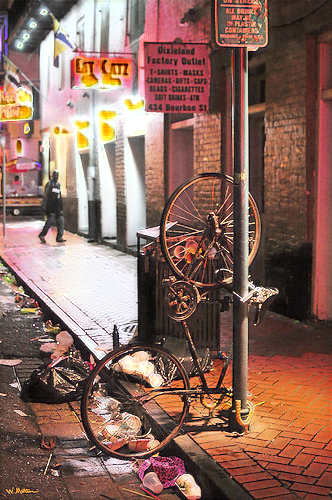}

    \includegraphics[width=.19\linewidth]{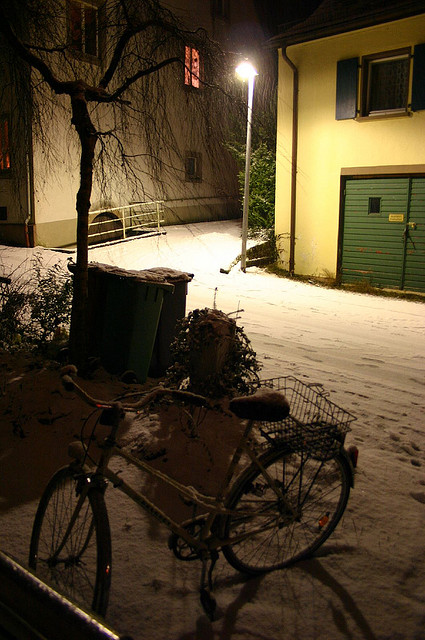}
    \includegraphics[width=.19\linewidth]{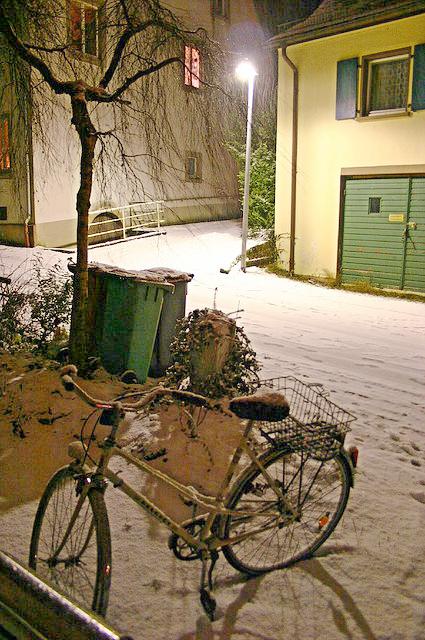}
    \includegraphics[width=.19\linewidth]{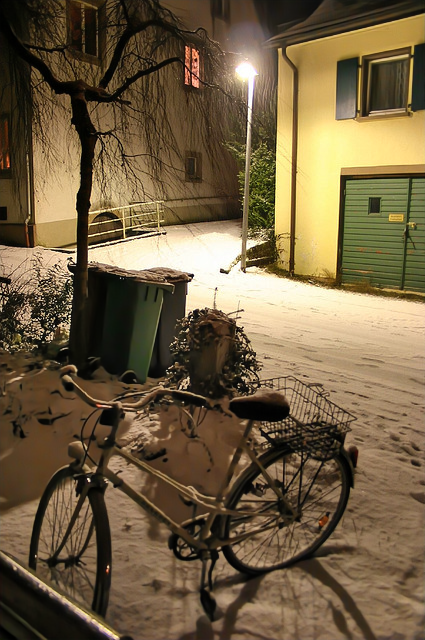}
    \includegraphics[width=.19\linewidth]{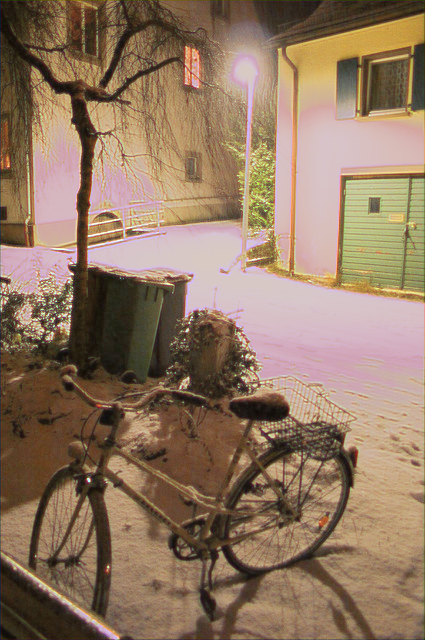}
    \includegraphics[width=.19\linewidth]{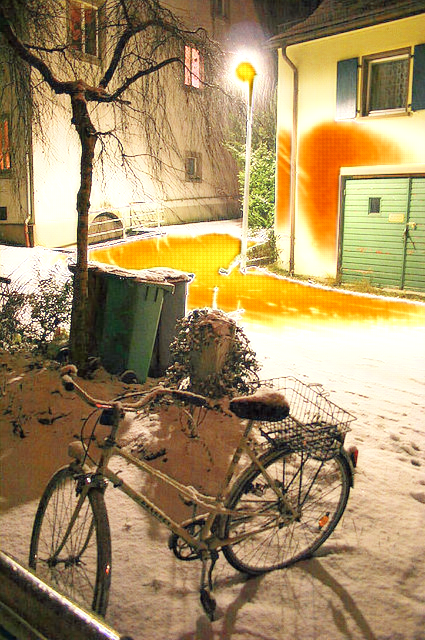}

    \includegraphics[width=.19\linewidth]{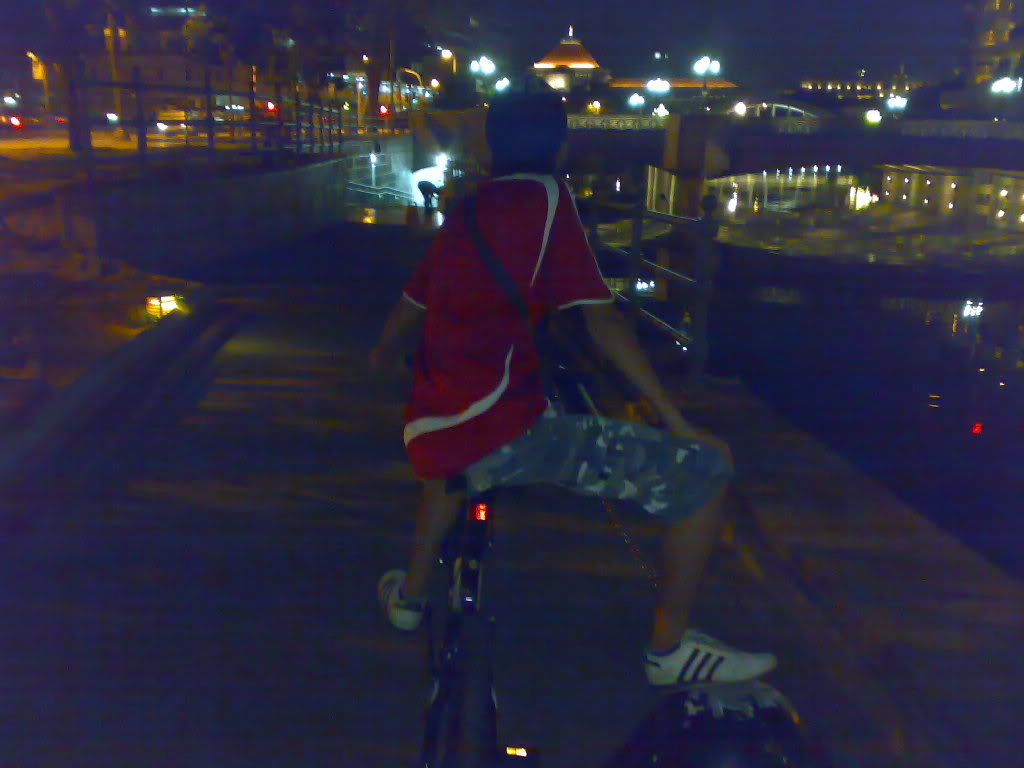}
    \includegraphics[width=.19\linewidth]{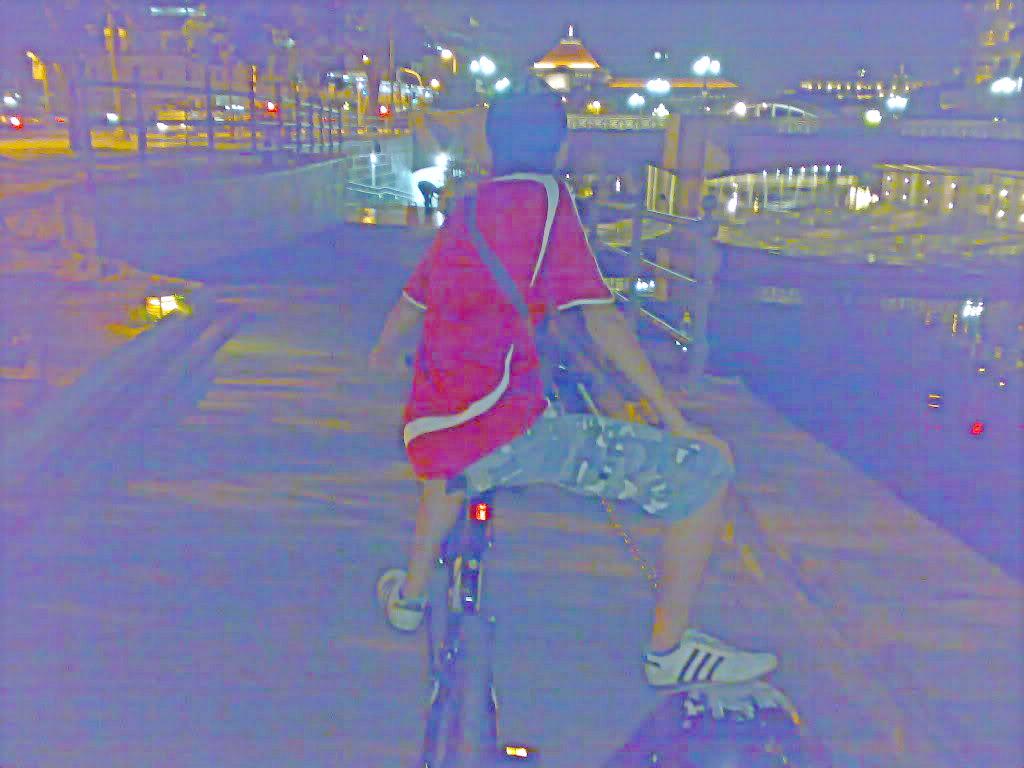}
    \includegraphics[width=.19\linewidth]{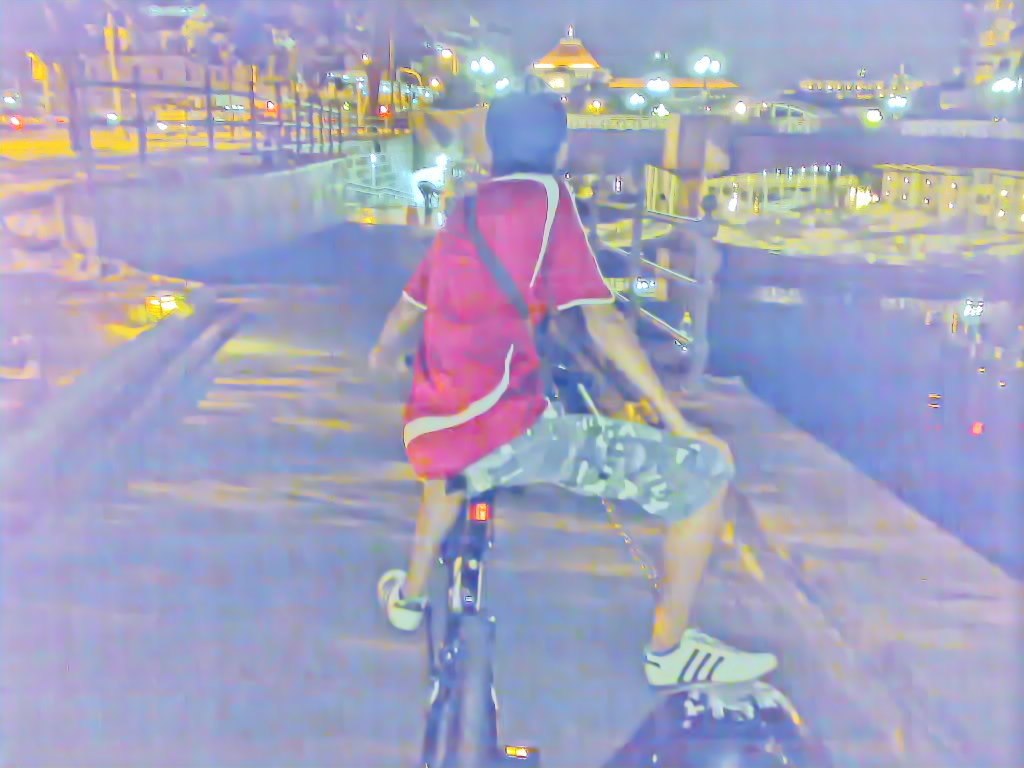}
    \includegraphics[width=.19\linewidth]{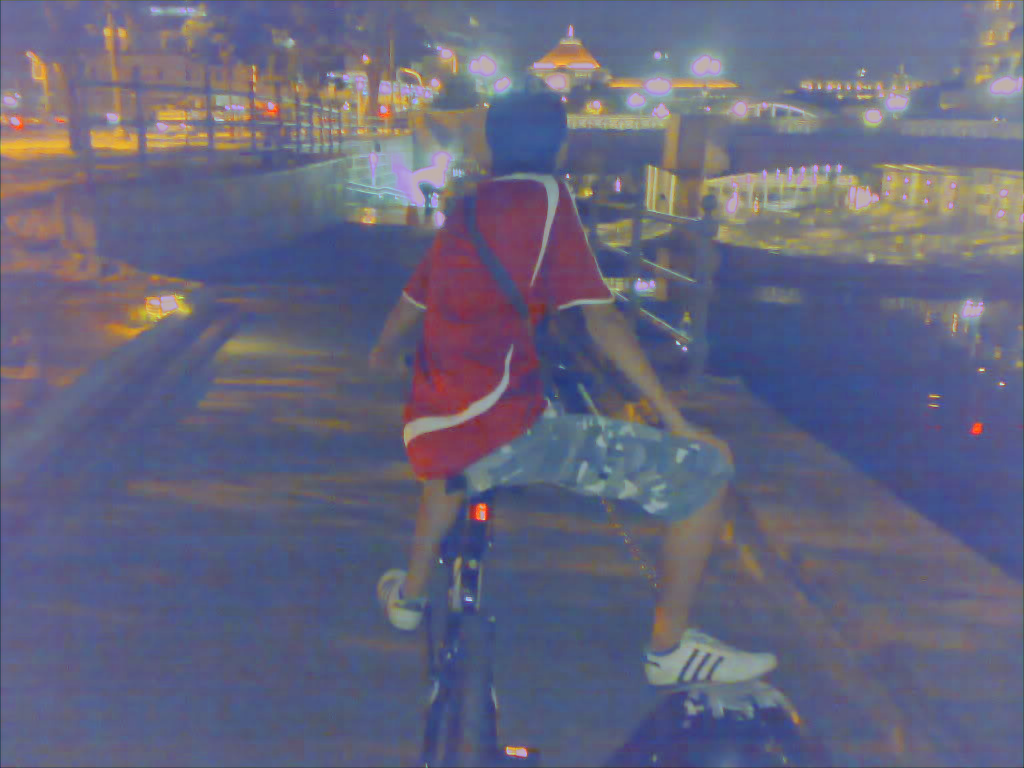}
    \includegraphics[width=.19\linewidth]{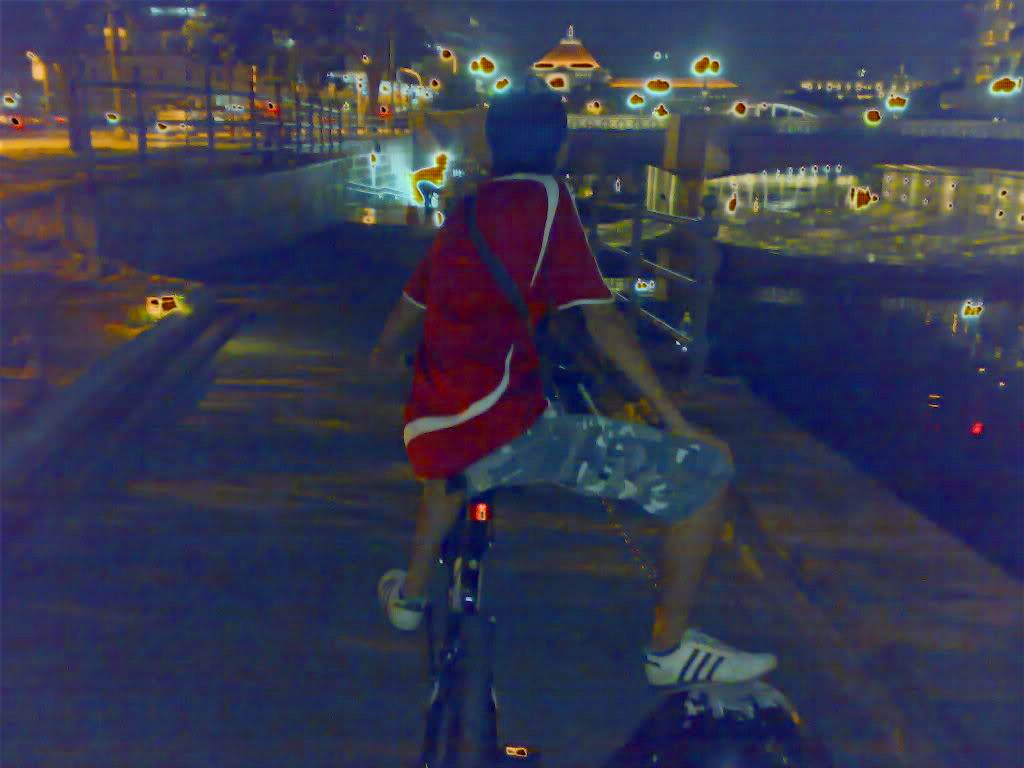}
    
    \begin{subfigure}{0.19\linewidth}
        \centering
        \subcaption{Input}
    \end{subfigure}
    \begin{subfigure}{0.19\linewidth}
        \centering
        \subcaption{Zero-DCE}
    \end{subfigure}
    \begin{subfigure}{0.19\linewidth}
        \centering
        \subcaption{Bread}
    \end{subfigure}
    \begin{subfigure}{0.19\linewidth}
        \centering
        \subcaption{IAT}
    \end{subfigure}
    \begin{subfigure}{0.19\linewidth}
        \centering
        \subcaption{RetinexFormer}
    \end{subfigure}
    \caption{Qualitative comparison on ExDark dataset. Please zoom in for more details. }
    \label{fig:quali_ExDark_1}
\end{figure}

In this section, we  first exhibited more comparison over existing methods on the ExDark dataset in Fig.~\ref{fig:quali_ExDark_1}. As can be found in these Bread~\citep{DBLP:journals/ijcv/GuoH23} presenting superior visual effects in most cases, the IAT~\citep{cui2022need} has the second-best performance where there exist artifacts in some cases. The ZeroDCE~\citep{Guo_2020_CVPR} has a good color restoration performance, but it suffers from unpleasant noise due to its non-denosing nature.  The outputs of RetinexFormer~\citep{cai2023retinexformer} have artifacts in multiple cases.

\begin{figure}[b]
    \centering
    \includegraphics[width=.32\linewidth]{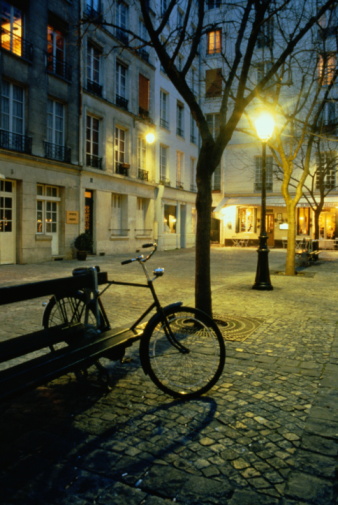}
    \includegraphics[width=.32\linewidth]{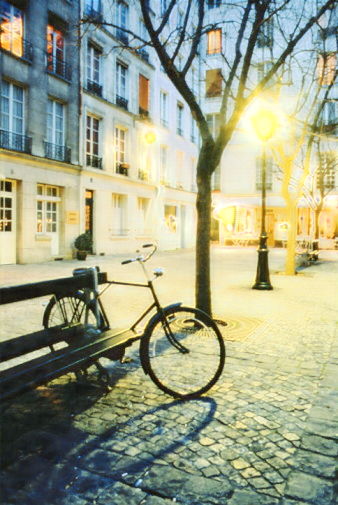}
    \includegraphics[width=.32\linewidth]{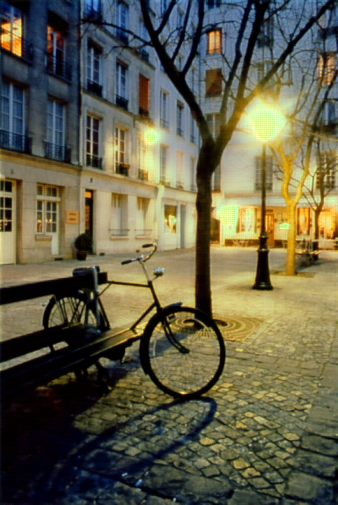}

    \includegraphics[width=.32\linewidth]{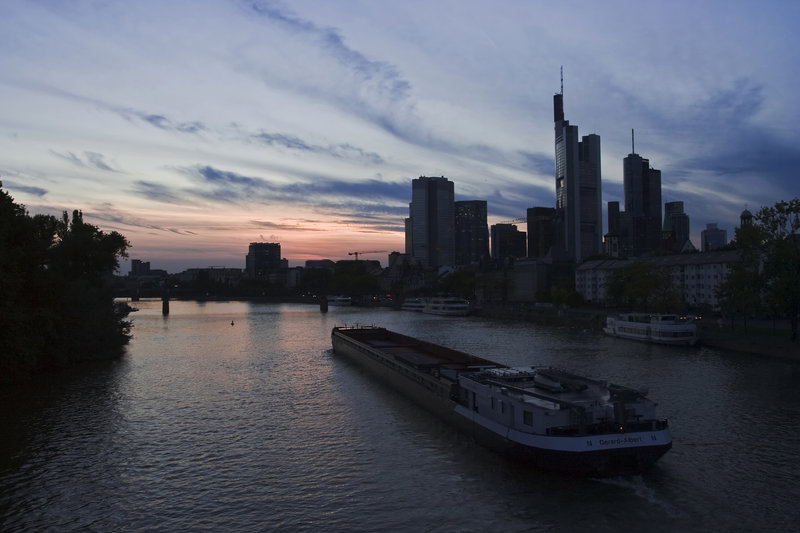}
    \includegraphics[width=.32\linewidth]{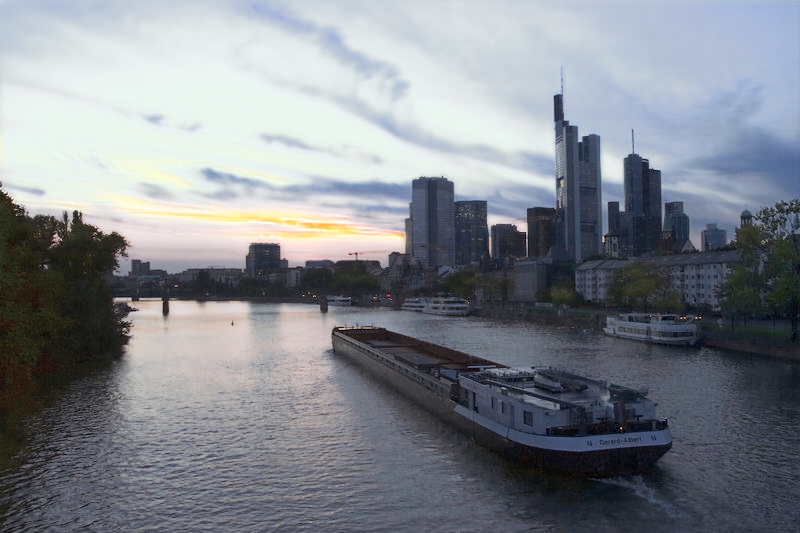}
    \includegraphics[width=.32\linewidth]{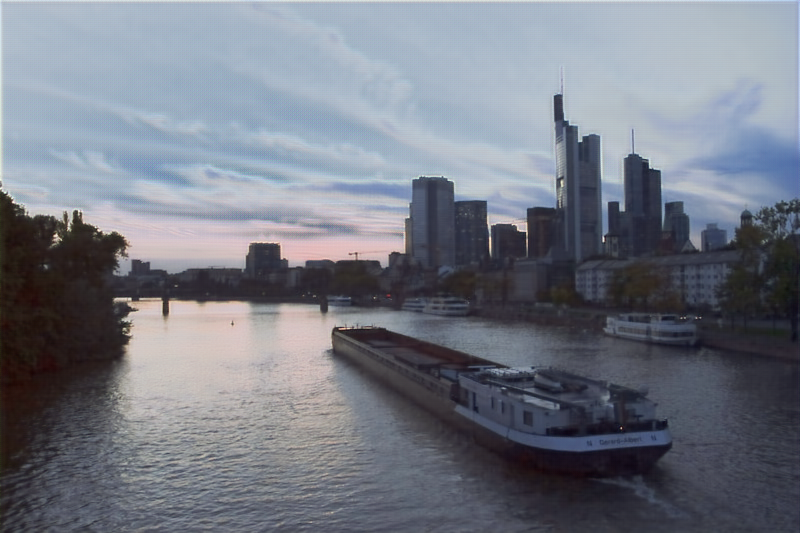}
    
    \begin{subfigure}{.32\linewidth}
        \centering
        \subcaption{Input}
    \end{subfigure}
    \begin{subfigure}{.32\linewidth}
        \centering
        \subcaption{RetinexFormer}
    \end{subfigure}
    \begin{subfigure}{.32\linewidth}
        \centering
        \subcaption{RetinexFormer + Ours}
    \end{subfigure}
    \caption{Qualitative comparison on ExDark dataset. Please zoom in for more details.}
    \label{fig:energy-pic-success}
\end{figure}

\begin{figure}[t]
    \centering
    \includegraphics[width=.32\linewidth]{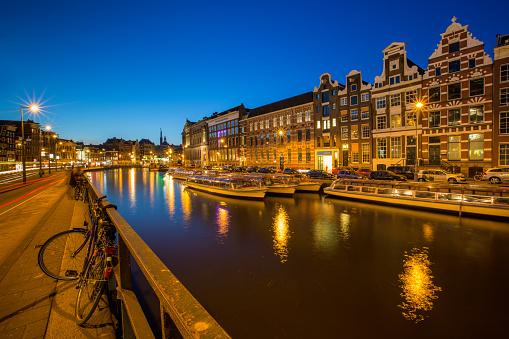}
    \includegraphics[width=.32\linewidth]{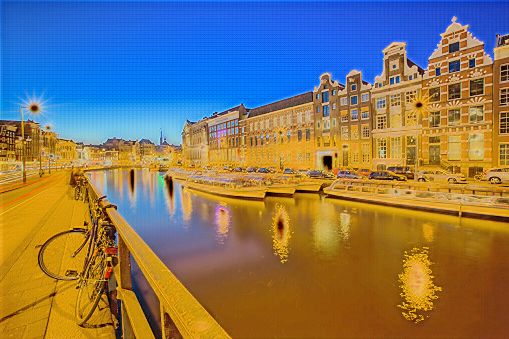}
    \includegraphics[width=.32\linewidth]{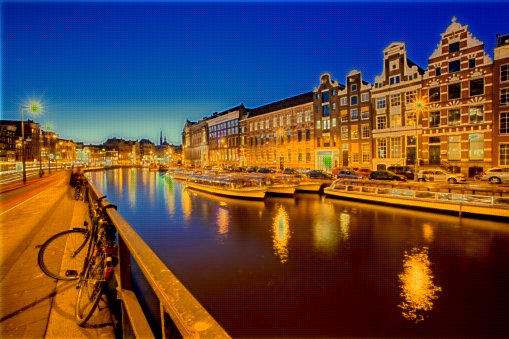}
    
    \begin{subfigure}{.32\linewidth}
        \centering
        \subcaption{Input}
    \end{subfigure}
    \begin{subfigure}{.32\linewidth}
        \centering
        \subcaption{RetinexFormer}
    \end{subfigure}
    \begin{subfigure}{.32\linewidth}
        \centering
        \subcaption{RetinexFormer + Ours}
    \end{subfigure}
    \caption{A Failure case on ExDark dataset. The model still failed to remove severe artifacts in some cases, even if our energy loss function is adopted.}
    \label{fig:failure_case}
\end{figure}

Qualitative results for models equipped with our energy function is presented in Fig.\ref{fig:energy-pic-success}. The model equipped with the energy function method can produce more natural outputs. However, as shown in Fig.\ref{fig:failure_case}, even equipped with our energy loss function, the model still failed to remove severe artifacts in some cases, including persistent checkerboard artifacts in the sky area, and the tendency for pixels in over-exposed areas to be out-of-bounds. Investigating this phenomenon and developing more sophisticated measures to alleviate it is out of the scope of this paper. Yet the case still demonstrates our ability to adjust images to a more natural exposure level.

\end{document}